\documentclass[12pt]{article}
\usepackage[a4paper, total={170mm, 230mm}]{geometry}

\usepackage[T1]{fontenc}
\usepackage[utf8]{inputenc}
\usepackage{newtxtext}
\usepackage{lmodern}
\usepackage{amsmath}
\usepackage{mathtools}
\usepackage{amssymb}
\usepackage{gensymb}
\usepackage{booktabs}

\usepackage[toc,page]{appendix}
\usepackage{lscape} 
\usepackage{caption}
\usepackage{subcaption}% 
\usepackage{booktabs}
\usepackage[hidelinks]{hyperref} % 
\usepackage{changepage}
\usepackage{float}
\usepackage{scalerel,amssymb}

\usepackage{setspace}
\usepackage{nomencl}
\makenomenclature
\usepackage[boxruled,linesnumbered]{algorithm2e}
\usepackage{hyperref}
\usepackage{etoolbox}
\usepackage{multicol}
\usepackage{sectsty}
\usepackage{blindtext}
\usepackage{graphicx}
\usepackage{capt-of}

\paragraphfont{\normalsize}
\renewcommand\nomgroup[1]{%
  \item[\bfseries
  \ifstrequal{#1}{R}{Roman Symbols}{%
  \ifstrequal{#1}{G}{Greek Symbols}{%
  \ifstrequal{#1}{A}{Acronyms}{}}}%
]}

\usepackage{capt-of}    
\usepackage{tabularx}   
\usepackage{blindtext}
\usepackage{enumitem}
\usepackage{tikz}
\usepackage{fdsymbol}
\usepackage{cite} 

\usetikzlibrary{calc,patterns,
                 decorations.pathmorphing,
                 decorations.markings}
\usetikzlibrary{positioning}
\usepackage[hang,flushmargin]{footmisc}

\title{Physics-informed transfer learning for SHM via feature selection}

\date{\vspace{-5ex}}
\begin{document}
\author{J.\ Poole\footnote{Corresponding author, email address: jpoole4@sheffield.ac.uk}, P.\ Gardner, A. J. Hughes, N.\ Dervilis, R. S. Mills, T. A. Dardeno \& K.\ Worden \\ Dynamics Research Group\\ School of Mechanical, Aerospace and Civil Engineering, University of Sheffield\\ Mappin Street, Sheffield S1 3JD, UK \\
	}
\maketitle

\pagenumbering{roman}
\pagenumbering{arabic}
\begin{abstract}
Data used for training structural health monitoring (SHM) systems are expensive and often impractical to obtain, particularly labelled data. Population-based SHM presents a potential solution to this issue by considering the available data across a population of structures. However, differences between structures will mean the training and testing distributions will differ; thus, conventional machine learning methods cannot be expected to generalise between structures. To address this issue, transfer learning (TL), can be used to leverage information across related domains. An important consideration is that the lack of labels in the target domain limits data-based metrics to quantifying the discrepancy between the marginal distributions. Thus, a prerequisite for the application of typical unsupervised TL methods is to identify suitable source structures (domains), and a set of features, for which the conditional distributions are related to the target structure. Generally, the selection of domains and features is reliant on domain expertise; however, for complex mechanisms, such as the influence of damage on the dynamic response of a structure, this task is not trivial. In this paper, knowledge of physics is leveraged to select more similar features, the modal assurance criterion (MAC) is used to quantify the correspondence between the modes of healthy structures. The MAC is shown to have high correspondence with a supervised metric that measures joint-distribution similarity, which is ultimately the primary indicator of whether a classifier will generalise between domains. The MAC is proposed as a physics-informed measure for selecting a set of features that behave consistently across domains when subjected to damage, i.e. features with invariance in the conditional distributions. When used in conjunction with established methods for aligning marginal distributions, the proposed approach yields transfers with high joint distribution similarities while remaining entirely unsupervised, thereby alleviating the need for costly labels. This approach is demonstrated on numerical and experimental case studies to verify its effectiveness in
various applications. \\

\emph{Keywords}: Transfer learning, SHM, physics-informed machine learning, domain adaptation
\end{abstract}
\section{Introduction}

%SHM is limited to PBSHM
In data-based SHM, diagnostics are often limited by the availability of data -- particularly labelled data -- as the collection of labelled data for engineering applications is often expensive and/or impractical. Therefore, the practical application of SHM is often limited to unsupervised machine learning, which has been shown to be capable of performing robust damage detection \cite{dervilis2014damage, FarrarC.R.CharlesR.2013Shm:}, but cannot provide contextual information such as damage location, type or extent \cite{FarrarC.R.CharlesR.2013Shm:}, which is necessary for important downstream tasks such as prognosis and decision-making \cite{hughes2021probabilistic}. Population-based SHM (PBSHM) is a field that aims to address this issue by increasing the potential information available by considering data from across a population of structures \cite{Bull2021, Gosliga2021, Gardner2021}. However, considering data from different structures will usually invalidate the assumption that the training and testing data were drawn from the same distribution -- an assumption made by conventional machine learning algorithms \cite{Murphy2014}.\\

This problem motivates the application of transfer learning (TL), a field of machine learning that aims to address issues of data scarcity by utilising labelled data from one or more source domains \cite{Zhuang2021}. TL algorithms typically use either limited target labels \cite{Yosinski2014, finn2017model} or unlabelled target data \cite{JialinPan2011, Long2013, Long2014, Long2015, Ho2002} to account for differences in the joint distributions.   The assumption is that the source domain (structure) is related to the target domain (another structure), so knowledge transfer can improve the performance in the target domain. Unsupervised TL, where no labels in the target domain are used \cite{pan2020transfer}, has gathered particular interest for SHM because obtaining labelled datasets for the target structure is often infeasible. \\

In unsupervised TL, the lack of labels means that measuring differences in the joint distributions directly is challenging; thus, these methods aim to minimise the discrepancy between the marginal distributions \cite{Ben-David2007, Gretton2012}, either by mapping the features into a shared space or by instance weighting \cite{Pan2010, Zhuang2021}. Therefore, a fundamental assumption is that the discrepancies between source and target joint distributions can be reduced by minimising a metric that measures marginal distribution discrepancy, which requires the conditional distributions of the original feature representation to be closely related. In the context of PBSHM, assuming vibration-based features are used, this assumption would allow the structures to have different absolute values of their natural frequencies but requires each structure have a similar response to a specific type and location of damage. Furthermore, for a given set of features, if the source and target domains do not satisfy this assumption, TL will likely lead to performance degradation -- so-called \emph{negative transfer }\cite{Zhang2020}. Many approaches to unsupervised TL assume that suitable domain and feature selection can be achieved via domain expertise \cite{Pan2010, Weiss2016, Zhuang2021}; however, in complex engineering systems this task is not trivial. Recently, principled methods to quantify structural similarity have been proposed; for example,  Gosliga \emph{et al.}\ have proposed quantifying physical similarity by considering abstract representations of structures as attributed graphs \cite{Gosliga2021}. However, these methods are yet to incorporate measured data and do not indicate which features will exhibit a similar response to damage (related conditional distributions).\\

To address these challenges this paper proposes using the modal assurance criterion (MAC) between a source and target structure \cite{allemang2003modal}, only utilising data from the undamaged state. In this way, the metric is informative of local discrepancies in the modal displacement between the structures; thus, is sensitive to conditional distribution shift relating to damage location. Moreover, by incorporating the MAC into a feature-selection criterion, an unsupervised TL approach based on physics is shown to select features that satisfy a fundamental assumption made in unsupervised TL -- that the features can be mapped into a shared feature space without using labelled target data. In addition, this approach only requires normal condition data in the target domain so can be applied before damage is observed in a target structure. To demonstrate the applicability of this methodology across different types of structures, results are presented from two distinct case studies: one involving the transfer of damage classifiers between structures within a large numerical population, and the other transferring a classifier between a population of two helicopter blades.\\

The paper is structured as follows. Section 2 summarises the related work, while Section 3 presents a discussion on TL and negative transfer in the context of PBSHM. Section 4 introduces the idea of using mode shapes as a means to quantify data similarity, and Section 5 investigates these the use of the MAC on a motivating example. Section 6 introduces the feature-selection criterion for physics-informed transfer, with results on a numerical population, suggesting this approach can significantly improve transfer when only a subset of features have related conditional distributions. Furthermore, Section 7 demonstrates the methodology's ability to transfer label information between experimental data from two heterogeneous helicopter blades. Finally, a discussion on physics-based similarity for TL in PBSHM is presented and conclusions are drawn. A list of acronyms used in this paper is also provided in Appendix A.\\

\section{Related work}
%literature 

To robustly transfer between structures, it is necessary to use a transfer pipeline that considers the similarity between the source and target domains, and their corresponding features, as well as the suitability of the chosen transfer learning algorithm \cite{Pan2010a}. This requires methods to select candidate source/target structures, feature extraction approaches to obtain transferable-damage sensitive features and transfer learning algorithms. Previous research has largely focused on the implementation of transfer learning algorithms \cite{wang2019characterizing}, selecting domains and features using engineering judgement. Meanwhile, this paper proposes a method to select transferable features as a method to enhance performance of transfer learning algorithms.\\

The application of transfer learning algorithms has been studied for several SHM tasks, including damage detection \cite{bull2021transfer, Michau2019}, localisation \cite{Gardner2020} and classification \cite{gardner2022population, dorafshan2018comparison}. These studies generally use fine-tuning or domain adaptation (DA) approaches. For example, fine-tuning allows for pre-trained models to initialise or set certain layers of the target model \cite{Yosinski2014}. These methods have been demonstrated for crack detection in images \cite{Gao2018, dorafshan2018comparison, zhu2020vision} and unprocessed frequency response data \cite{cao2018preprocessing, teng2023structural}. While fine-tuning can reduce the need for large datasets, it is limited in sparse data scenarios as it still requires representative labelled data.\\

Domain adaptation has received significant attention as a means of addressing scenarios with sparse target data. In a PBSHM setting, Gardner \emph{et al.} have shown that DA can be used to transfer localisation labels between numerical and experimental structures \cite{Gardner2020}, two heterogeneous aircraft wings \cite{gardner2022population}, and between pre- and post-repair states in aircraft wings \cite{Gardner2020b}. In addition, Bull \emph{et al.} used a population of six experimental tailplanes to demonstrate transferring a damage detector \cite{bull2021transfer} and Figueiredo \emph{et al.} \cite{figueiredo2023transfer} applied a similar approach for damage detection in bridges. There have also been a number of applications of deep-DA architectures proposed to perform fault diagnosis in machines under changing loading conditions and rotation speeds \cite{wang2020triplet, Li2019, Michau2019, Li2020, jiao2020residual}.\\

While these previous works have demonstrated the potential for transfer learning in various tasks, most case studies presented relied on engineering judgement to select transferable-damage sensitive features. In practice, this task may be challenging in complex structures. Another approaches uses marginal-distribution distance measures for domain selection \cite{gardner2022population}. However, as will be shown in Section 4, to guarantee a predictive model will generalise well across domains it is important to measure the joint distribution discrepancy — highlighting a major limitation of previously used data-driven approaches. As far as the authors are aware, the only other approach investigating mode shape similarity in the context of transfer learning is that of \cite{de2025similarity}, which used cosine similarity to select features from a potential set of natural frequency and mode shape features. The approach proposed in this paper differs fundamentally from previous work, as it does not require mode shapes from multiple health states — either for computing similarities or for use as features. Instead, it relies only on a small sample of mode shapes from the undamaged structure to guide the transfer learning process. A numerical study demonstrates that the similarity of undamaged mode shapes can serve as a proxy for joint-distribution similarity between domains. Furthermore, initial results from this paper were previously presented in two earlier conference papers \cite{poole2023physics, poole2023towards}. In addition, as far as the authors are aware only one previous work has incorporated engineering knowledge directly into a transfer learning pipeline \cite{Xu2020}, where building height was used to weight source domains in a multi-source DA algorithms. This differs to the objective of this paper, which identifies a physics-based measure for indicating joint distribution similarity without labelled target data and aims to select transferable features prior to applying transfer learning algorithms.  \\

This paper demonstrates that the modes associated with the undamaged structure can serve as indicators of joint-distribution similarity. A widely used modal analysis tool, the MAC \cite{allemang2003modal}, is then applied to inform the selection of features for use with transfer learning. The key contributions are  summarised as follows:

\begin{itemize}
    \item Similarity between the mode shapes of only the undamaged structure are demonstrated to be informative of structural differences that influence whether a source classifier can be generalised to target test data using DA. Use of mode-shape similarity addresses limitations of previous approaches that only have access to unlabelled target data, as these methods only measure marginal distribution distance, whereas generalisation of data-based models rely on joint-distribution similarity.
    \item A method based on a popular measure for comparing mode shapes, the MAC \cite{allemang2003modal}, is formulated into a feature selection criterion, providing a principled method to select features with high correspondence in the joint-distributions. 
    \item The feature selection criterion was demonstrated with DA to present a complete transfer learning pipeline. Thus, this study presents one of the first transfer learning pipelines to allow for the generalisation of a multi-class classifier in SHM, with validation presented using a novel laboratory dataset relating to different helicopter blades. 
\end{itemize}
\

 By presenting a transfer learning pipeline for SHM, this paper also represents a shift in previous transfer learning literature. Previous literature has mostly considered feature selection to deal with dimensionality issues \cite{wen2019empirical, souza2021investigation}, or to select sets of features with the lowest marginal distribution distance \cite{uguroglu2011feature, persello2015kernel, chen2019cross,nguyen2018particle, sanodiya2020particle}. However, in many engineering scenarios, these methods may not be sufficiently robust - as discussed in Section 4 -- and they ignore engineering knowledge that could be used to measure similarity.  As such, this paper aims to incorporate modal information into a feature-selection criterion to address limitations with a purely data-based approach for measuring domain similarity. \\

\section{Transfer learning and the problem of negative transfer}

Information sharing between structures has the potential to reduce the cost, and increase the reliability of monitoring systems. However, differences in the material properties, boundary conditions and geometry between the structures will lead to changes in the data distributions \cite{Gosliga2021}. One method to facilitate information sharing under distribution shift is transfer learning. The definition of TL requires two objects to be defined: 

\begin{itemize}
    \item A \emph{domain} $\mathcal{D}=\{\mathcal{X}, p(\mathbf{x})\}$, defined by a feature space $\mathcal{X}$ and a marginal probability distribution $p(\mathbf{x})$ on that space.
    \item A \emph{task} for a given domain is defined by $\mathcal{T}=\{\mathcal{Y},f(\cdot)\}$, where $\mathcal{Y}$ is the label space and $f(\cdot)$ is a predictive function learnt from a finite sample  $\{\mathbf{x}_{i},y_{i}\}_{i=1}^{n}$, where $\mathbf{x}_i \in \mathcal{X}$ and $y_i \in \mathcal{Y}$. The predictive function can also be viewed as modelling the conditional distribution $p(y| \mathbf{x})$.
\end{itemize}

This paper primarily considers unsupervised TL to allow for knowledge transfer without labels relating damage states and extreme EoVs \cite{pan2020transfer} \footnote{It should be noted that this paper is still applicable to supervised TL, as similarity of a given feature space is a core assumption of all TL.}. In \emph{unsupervised TL}, a source domain $\mathcal{D}_s=\{\mathbf{x}_{s,i},y_{s,i}\}_{i=1}^{n_s}$, with $n_s$ source instances $\mathbf{x}_{s,i}$, each with labels $y_{s,i}$ and a target domain $\mathcal{D}_t=\{\mathbf{x}_{t,j}\}_{j=1}^{n_t}$ with $n_t$ unlabelled target instances, $\mathbf{x}_{t,j}$, are used to learn a predictive function that generalises to the target domain\footnote{This definition of TL follows \cite{pan2020transfer}, unsupervised TL is also often referred to as transductive TL as in \cite{Pan2010}.}.  Commonly, unsupervised TL assumes that there is covariate shift \cite{Zhuang2021}, i.e. $p_s(\mathbf{x}) \neq p_t(\mathbf{x})$ and $p_s(y|\mathbf{x}) = p_t(y|\mathbf{x})$. This assumption is often too restrictive, motivating mapping-based approaches to unsupervised DA, which instead assumes $p_s(\mathbf{x}) \neq p_t(\mathbf{x})$ and $p_s(y|\mathbf{x}) \neq p_t(y|\mathbf{x})$, and a mapping $\phi$ exists, that projects the domains into a shared feature space, where $p_s(\phi(\mathbf{x})) = p_t(\phi(\mathbf{x}))$ and $p_s(y|\phi(\mathbf{x})) = p_t(y|\phi(\mathbf{x}))$. In unsupervised DA, the lack of labels makes estimating (and directly minimising) the discrepancy between the conditional distributions challenging; thus, unsupervised measures that measure marginal distribution discrepancy are leveraged to learn $\phi$ \cite{Zhuang2021}. As such, DA requires a strong correspondence between the conditional distributions in the source and target domains. \\

\begin{figure}[h!]
\centering
\includegraphics[width=12cm]{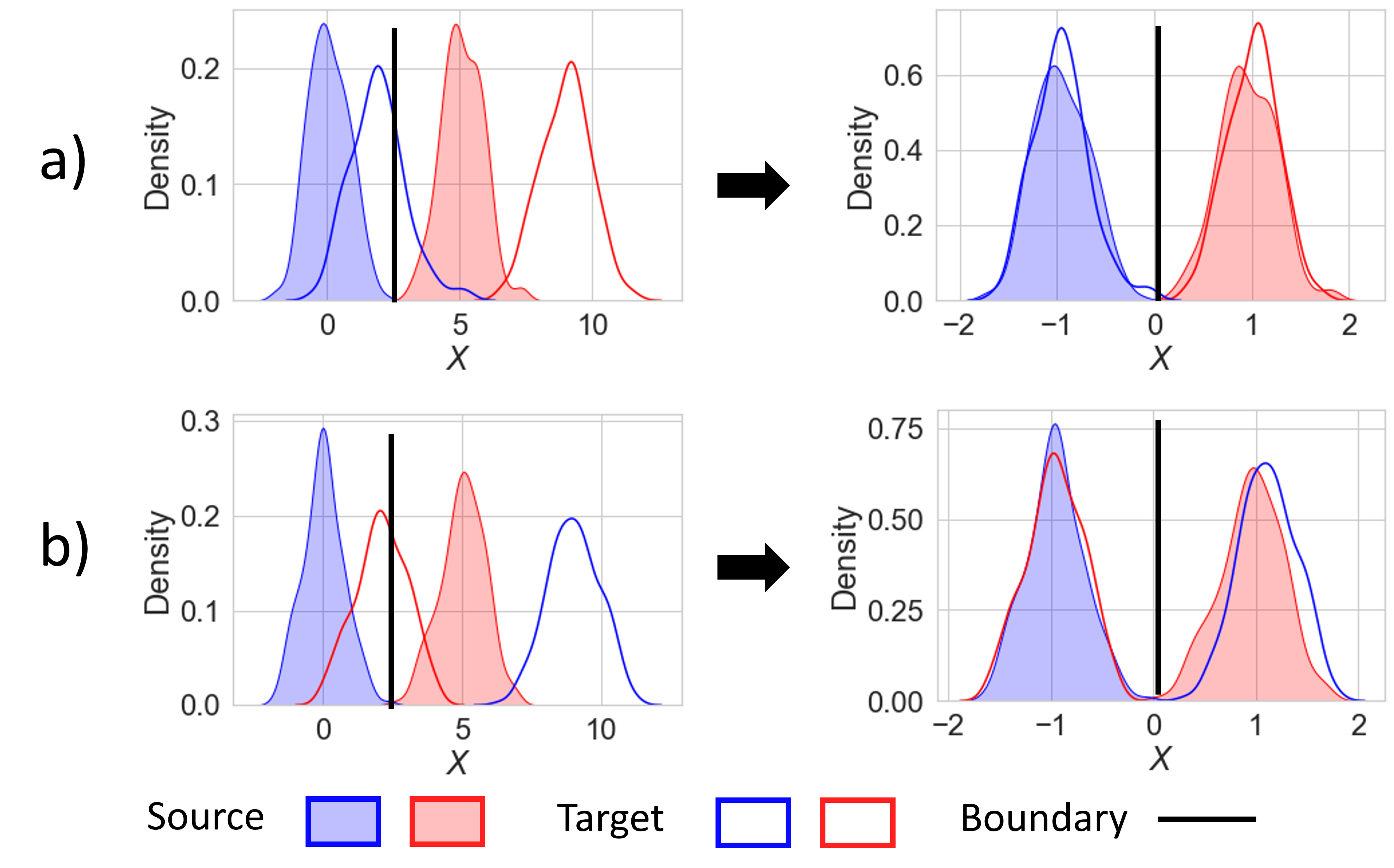}
\caption{A demonstration of positive transfer by transferring a binary classifier when domains have similar conditional distributions, presented in (a), and the potential of negative transfer when conditional distributions are unrelated is shown in (b).}
\label{fig:condproblem}
\end{figure}

Determining when minimising the distance between the marginal distributions alone is sufficient is not trivial; if this assumption is not valid, DA will result in performance degradation -- referred to as negative transfer \cite{wang2019characterizing} -- which may have critical consequences in SHM, where poor generalisation of a source classifier can lead to unnecessary inspections or to severe damage by missing critical maintenance. Furthermore, the lack of labels makes it difficult to assess performance using traditional validation techniques, such as cross-validation, and certain issues, such as label switching, mean instances of negative transfer may be indiscernible from successful transfer. To illustrate this issue, Figure \ref{fig:condproblem} shows a toy example of a shift in the distributions, which can be reduced using DA, as shown in the right panel. However, without labels or prior knowledge, it is challenging to distinguish between cases where the conditional distributions are related, meaning a classifier would generalise well after DA as shown in Figure \ref{fig:condproblem}a, and the case where the labels are flipped, as shown in Figure \ref{fig:condproblem}b. Minimising the marginal distribution distance in the latter case would result in a model that misclassifies the target.  \\

%conditional distribution figure

This problem highlights the need for robust methods to select features (and domains), such that unsupervised DA is unlikely to result in negative transfer. In unsupervised TL, There are two main approaches to select domains and corresponding features -- using unsupervised measures or domain expertise. As previously mentioned, whilst domain expertise is valuable, selecting structures and sets of features that satisfy transfer conditions using domain expertise alone is not trivial. The previous TL literature has also utilised several marginal distribution distance measures \cite{Chen2010, Dai2007, Si2010, Wang2011}. Recent TL research has largely focused on two divergence measures \cite{Zhuang2021}, the \emph{maximum mean discrepancy} (MMD) \cite{Gretton2012a} and the \emph{proxy-$\mathcal{A}$ distance} (PAD) \cite{Ben-David2007}.  However, these measures have critical several limitations for PBSHM applications:
\begin{itemize}
    \item In an unsupervised transfer learning setting (where there are no target labels), these measures are limited to estimating marginal distribution discrepancy, which will not provide a robust measure of similarity if there are differences in the conditional distributions. While there are some attempts to estimate the conditional-distribution discrepancy by using pseudo-labels \cite{Long2014}, the pseudo-labels themselves will only be accurate if the initial conditional-distribution discrepancy is small.
    \item These measures may require large datasets; for example, the MMD relies on data being sufficient to estimate a mean in high-dimensional feature spaces, and the PAD may require sufficient data to train a classifier with many parameters, such that it has sufficient complexity to be a valid measure on the distributions.
    \item The available data in the target domain may only represent a small subset of the possible classes; for example, at the start of a monitoring campaign of a target structure, only normal-condition data would be available, whereas the source might have a range of health-states to transfer. In such cases, the source and target label space would be a subset of the source label space ($\mathcal{Y}_t \subset \mathcal{Y}_s$), meaning the unsupervised measures would not indicate whether the underlying distributions differ, but rather that the available subset of the distributions differs.
\end{itemize}

The issue of negative transfer and the challenges in validating models, motivate the need for robust methods for selecting domains, and suitable sets of features within these domains. This paper focuses on developing a methodology to address the latter problem; however, this measure could also be used to aid domain selection \cite{hughes2023quantifying}. To this end, the following section presents a physics-based measure to address the aforementioned challenges encountered when only considering unsupervised data-based similarity measures. The core motivation is that for a subset of features that relate to similar physical mechanisms, unsupervised DA should lead to positive transfer, as discussed in \cite{rojas2018invariant}.

\section{Physics-based similarity}

In the context of SHM, additional insight can be gained by exploiting physical knowledge between structures.  As outlined in Rytters hierarchy \cite{Rytter1993}, assuming damage detection can be achieved using unsupervised approaches, the next task an SHM system should attempt is damage localisation. To perform damage localisation by leveraging information from a source domain, the influence of damage in a specific location $l$, on the response of the structures should be similar. Specifically, using only the limited available target data, it must be possible to find a mapping such that $p_s(l|\phi(\mathbf{x})) = p_t(l|\phi(\mathbf{x}))$. For simplicity, the remainder of the paper will refer to damage location as a discrete location, denoted by  labels $y$. \\

Assuming linear behaviour, it is well established that the modal parameters completely characterise the dynamic response of a structure \cite{FarrarC.R.CharlesR.2013Shm:}. These properties have been shown to be sensitive to local changes in stiffness caused by damage \cite{worden2004overview, farrar2000structural, kim1992applications, farrar1996damage}. This paper leverages this relationship with local stiffness in a different way, using the mode shapes to assess structural similarity in the context of transfer. Specifically, it is proposed that the similarity of mode shapes  relating to the undamaged structure be used to determine which frequency-based features are suitable for transfer learning. In this paper, similarity is assessed using the MAC, a widely used tool for comparing mode shapes \cite{allemang2003modal}.\\
 
    Since the objective is to facilitate transfer in scenarios where data in the target domain are sparse, it is proposed that similarity assessment can be performed only using mode shapes derived from the undamaged structure (normal-condition data). The general idea is that the mode shapes indicate areas of high strain for a given mode; thus, the locations where a given mode will be sensitive to damage. For example, the locations of the nodes will indicate the regions of a structure where damage will have the greatest influence on the modal features, as the stiffness in this region has the largest influence on the modal response. As such, assuming that the nodal patterns of the mode shapes are similar between systems, it is hypothesised that vibration-based features (e.g.\ natural frequencies)will have a similar sensitivity to damage, thus meaning transfer should be possible. It should be emphasised that the approach presented in this paper only requires mode shapes from a limited time period where the structure is undamaged. As such the high cost associated with obtaining a set of mode shapes -- i.e.\ the requirement for a dense sensor network and the need for experts to perform modal analysis -- could be reduced by performing a one-time analysis.\\
    
    An illustrative example of the relationship between the influence of damage and the nodal pattern of a mode is presented in Appendix B. Additionally, a more in-depth justification that the damage response between two structures will be similar for a given mode, given that the mode shapes of the normal conditions are related, is presented in Appendix C. \\

A widely used tool for the comparison of mode shapes is the MAC \cite{allemang2003modal}. The MAC has been used for comparison of modes both within and between structures, and it is widely used for model updating \cite{allemang2003modal}; however, its use has not been investigated in relation to transfer learning. The MAC is a normalised scalar product between each pair of modal vectors $\boldsymbol{\psi}^{(i)}_{s}$ and  $\boldsymbol{\psi}^{(j)}_{t}$ from two modal matrices $\Psi_s$ and  $\Psi_t$, which in this paper relate to the source and target domains respectively. The scalars are then arranged into a MAC matrix, assuming real-valued modal vectors; it is given by,
\begin{equation}
    \text{MAC}(i,j) = \frac{|\boldsymbol{\psi}^{(i)T}_{s}\boldsymbol{\psi}^{(j)}_{t}|^2}{\boldsymbol{\psi}^{(i)T}_{s}\boldsymbol{\psi}^{(i)}_{s}\boldsymbol{\psi}^{(j)T}_{t}\boldsymbol{\psi}^{(j)}_{t}}
\end{equation}
where $\text{MAC}(i,j)\in [0,1]$, with $0$ indicating no correspondence and $1$ is complete correspondence. If both modal matrices correspond to similar modes, the leading diagonal will be close to unity. Here the MAC is compressed into a scalar; a measure can then be given by,
\begin{equation}
    d_{\text{MAC}}(\Psi_s,\Psi_t) = \frac{1}{d}\sum_{i,j \in \mathcal{I}} \text{MAC}(i,j)
\end{equation}
where $\mathcal{I} = { (\mathbf{v}_{s}, \mathbf{v}_t) \mid \mathbf{v}_{s}, \mathbf{v}_t \in \mathbb{R}^d}$ is the pairs of feature indices, where $\mathbf{v}_{s}, \mathbf{v}_t$ are vectors of integers representing the source and target indices corresponding to the features being compared respectively and $d$ is the total number of features being compared in each domain, so the measure is normalised $d_{MAC}\in [0,1]$; this measure is called the MAC-discrepancy. \\

\section{Motivating case study: evaluation of similarity measures}

Measures that quantify the similarity between domains are central to any unsupervised TL approach. Measures that do not require label information are of particular importance in unsupervised DA, as they typically form the basis of the cost function used to learn shared features. A pitfall of unsupervised DA is that it only leads to improved generalisation under the condition that the conditional distributions are sufficiently related between domains. In addition, ensuring this condition is satisfied is challenging without using label information. This section presents results for attempting transfer, using unsupervised DA, across a numerical case study. It is shown that the MAC-discrepancy is correlated with the accuracy resulting from transfer; thus, motivating its use in conjunction with DA to ensure features used for transfer have sufficiently related conditional distributions; the methodology for feature selection using the MAC is presented in the proceeding section.
 
\subsection{Numerical population: a classic SHM example}

\begin{figure}[b!]
    \centering
    \includegraphics[width=\textwidth]{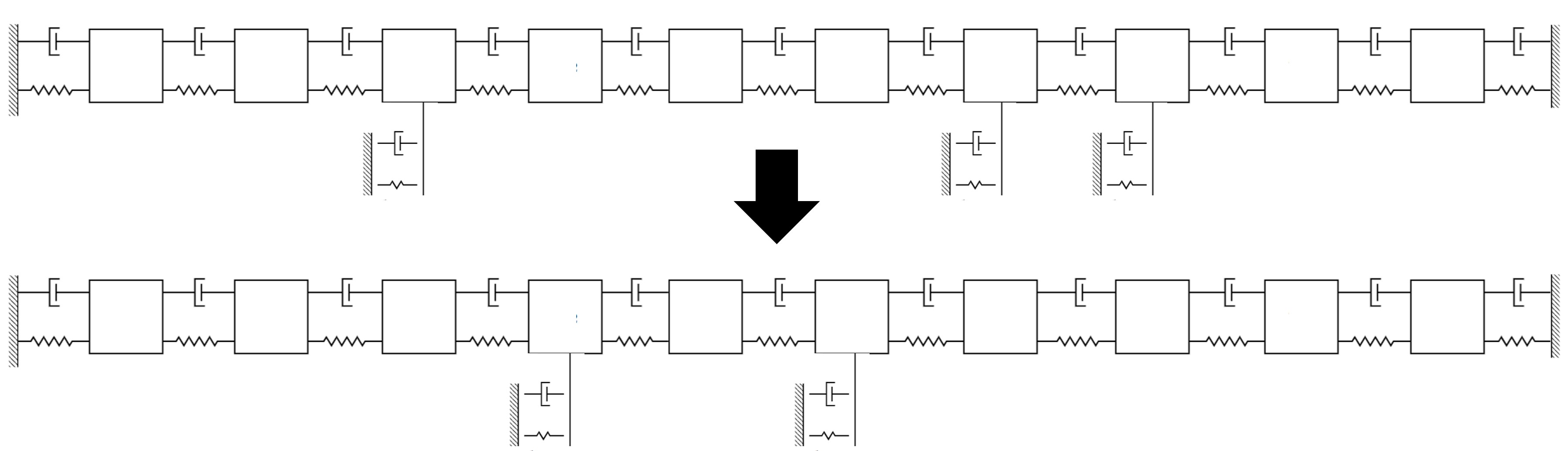}
    \caption{An example pair of numerical structures from the numerical case study.}
    \label{fig:ndof}
\end{figure}

The population presented in this section considers a number of challenging transfer problems, where structures differ by the number, and location of connections to ground.  The population was generated via the lumped-mass approach to simulate 20 structures, with ten degrees of freedom (DoF). This paper considers homogeneous TL, where the label space is equivalent, i.e. $\mathcal{Y}_s=\mathcal{Y}_t$. Thus, connections between masses were kept consistent and variation was introduced by randomly adding additional connections to the ground, an example of two structures is shown in Figure \ref{fig:ndof}. Between one and three connections were added to the ten DoF chain of masses; these were added in random locations drawn independently for each structure. The central six masses were considered as candidate locations for extra connections. \\

Geometries for each structure were the same; geometries that the masses and stiffness were derived from, as well as details of the assumed material variation are presented in Table \ref{tab:prob}. Each lumped mass was assumed to be a rectangular volume, parameterised by a length $l_m$, width $w_m$, thickness $t_m$, and density $\rho$, with the density sampled from a Gaussian distribution to represent manufacturing variation. The stiffness between masses was assumed to follow the model for a cantilever beam, where stiffness was found as the tip stiffness of a cantilever beam, $k_b=\frac{3EI}{l_m^3}$; the same geometry as the mass was used to model the beam.  The elastic modulus $E$ was also drawn from a Gaussian distribution for each sample to introduce variability. The stiffness for the ground connections was modelled following the same approach. Damping $c$ was not derived from a physical model, instead, it was drawn from a Gamma distribution directly. Damage at a given degree of freedom was modelled as an open crack, located in the centre of each connection, with a length of 0.1m. It was modelled as a reduction in stiffness following the model presented by Christides and Barr \cite{Christides1984}. \\

\begin{table}[h!]
\centering
\begin{tabular}{llll} 
\hline
\multicolumn{1}{l}{}               & Unit     & Values                                           \\ 
\hline
Beam geometry~$\{l_b,w_b,t_b\}$     & $m$     & $\{5.6,1.1,6\}$                         \\ 
Mass geometry~$\{l_m,w_m,t_m\}$     & $m$     & $\{5.6,1.1,6\}$                    \\ 
Crack geometry~$\{l_{cr},l_{loc}\}$ & $m$     & $\{0.1,2.8\}$                            \\ 
Elastic modulus $E$                 & $GPa$    & $\mathcal{N}(20,1\times 10^{-9})$ \\ 
Density~$\rho$                      & $kg/m^3$ & $\mathcal{N}(2300,20)$                         \\ 
Damping coefficient~$c$             & $Ns/m$   & $\mathcal{G}(8,0.8)$            \\
\hline
\end{tabular}
\caption{ Properties for the numerical case study.}
\label{tab:prob}

\end{table}

Having obtained the parameters for the 20 systems, the damped natural frequencies $\mathbf{\omega}_{d}$ and modes $\Phi$ were calculated by solving the eigenvalue problem. Ten classes were generated, corresponding to no damage and damage on the central nine springs. A total of 100 samples were generated for each damage-state, giving 1000 samples in total, which were evenly divided into training and testing datasets. Each pair of structures that did not have identical or symmetric ground connections was considered a transfer task, giving 360 transfer tasks. For each task, damage classification was attempted, considering the normal condition and nine damage locations, pertaining to damage to all nine springs between masses (not including any ground connections). The natural frequencies were used as features $X \in \mathbb{R}^{n \times 10}$, and the mode shapes of the normal conditions were utilised in the similarity measure. For each task, the source structure was assumed to be labelled and only normal-condition data in the target were assumed to include any labels. \\

\subsection{Transfer learning methodology}

To assess whether the MAC can be used to quantify conditional distribution similarity, first this section establishes whether the MAC can indicate when unsupervised DA will result in low joint distribution discrepancy. To this end, several measures are used to quantify discrepancy between the source and target after applying DA for each of the tasks in the numerical population.\\

The DA technique here was \emph{normal condition alignment} (NCA), which applies a set of linear transformations that align the normal conditions \cite{poole2022statistic}. NCA aims to reduce the risk of aligning data generated via different processes, by assuming that data gathered at the start of the operation of a structure were generated by the ``normal-condition" -- a common assumption made for novelty detection \cite{Dervilis2015}; thus, this method could be used with the MAC-discrepancy when only normal-condition data are available in the target domain, which will often be the case in PBSHM.\\

This method first standardises the source domain, and the normal conditions are then aligned by, 
\begin{equation}   \boldsymbol{z}_{t, i}=\left(\frac{\boldsymbol{x}_{t, i}-\boldsymbol{\mu}_{t,n}}{\boldsymbol{\sigma}_{t,n}} \right)\boldsymbol{\sigma}_{s,n}+\boldsymbol{\mu}_{s,n}
\end{equation}
\noindent where $\boldsymbol{z}_{t, i}$ is the transformed sample, while $\boldsymbol{\mu}_{s,n}$, $\boldsymbol{\mu}_{t,n}$ and $\boldsymbol{\sigma}_{s,n}$, $\boldsymbol{\sigma}_{t,n}$ are the means and standard deviations of the normal-condition data for the source and target respectively.

NCA was selected as a DA method here as it accounts for differences in absolute position and scale. As such, if the features have similar sensitivity to damage in a given location (i.e. the proportion that the frequency changes), a classifier trained on the source domain should generalise to the target after NCA. \\ 

Following NCA, a K-nearest neighbours classifier (KNN), with one nearest neighbour was trained on the source domain, which is commonly used to assess DA algorithms \cite{Zhuang2021}.  Here a KNN is chosen since, if the distributions of the domains are well aligned, the data should be close in Euclidean space. \\

The MAC-discrepancy was evaluated for each transfer task to investigate whether it is capable of indicating when the source and target domains were related enough to perform unsupervised DA. In addition, two unsupervised measures, the MMD and the PAD, were used to measure the marginal distribution discrepancy; for more details on these measures the interested reader may refer to \cite{Ben-David2007, Gretton2012a}.  These measures were chosen for their prevalence in DA and were implemented to verify whether a purely data-driven approach could effectively determine when unsupervised DA can be successfully applied. A fully supervised divergence measure, the joint-MMD (JMMD) \cite{Long2014}, was also used to show that joint distribution discrepancy is the primary indicator of transfer robustness. The JMMD is a modification of the MMD \cite{Long2014}, which is a sum of the MMD between the marginal and class-conditional distributions. It should be noted this measure is not applicable in practice as it requires labels in the target domain. Note that a lower value of these data-based measures indicates greater similarity. The objective of this comparison is to understand whether the MAC can provide additional information compared to using unsupervised data-based measures alone, to motivate its application for feature selection prior to the application of DA to facilitate more robust transfer.\\

Before applying the data-based measures, several parameters must be set. The PAD requires the specification of a suitably complex classificatier to discriminate between domains; in this paper, a support vector machine (SVM) with an radial basis function (RBF) kernel was utilised. The SVM was trained using 70\% of the test data and the PAD was calculated using the error of the remaining 30\% of the test data. The MMD projects data into an RKHS via a universal kernel; here an RBF kernel was used and the length scale was specified as the median of the pairwise distances between domains (the median heuristic), a common heuristic for unsupervised hyperparameter selection in DA \cite{Gretton2012a, JialinPan2011, Long2015, Gardner2020}. The MAC-discrepancy was calculated using the mode shapes obtained from a single observation of the undamaged structure. The accuracy of the test data was used as a classification quality measure; accuracy is given by,  
\begin{equation}
    \text{Accuracy} = \frac{TP + TN}{TP + TN + FP + FN}
\end{equation}
\noindent where $TP$ is the number of true positives, $TN$ is the number of true negatives, $FP$ is the number of false positives, and $FN$ is the number of false negatives.\\

\subsection{Results}

\begin{figure}[h!]
\centering
\includegraphics[width=12cm]{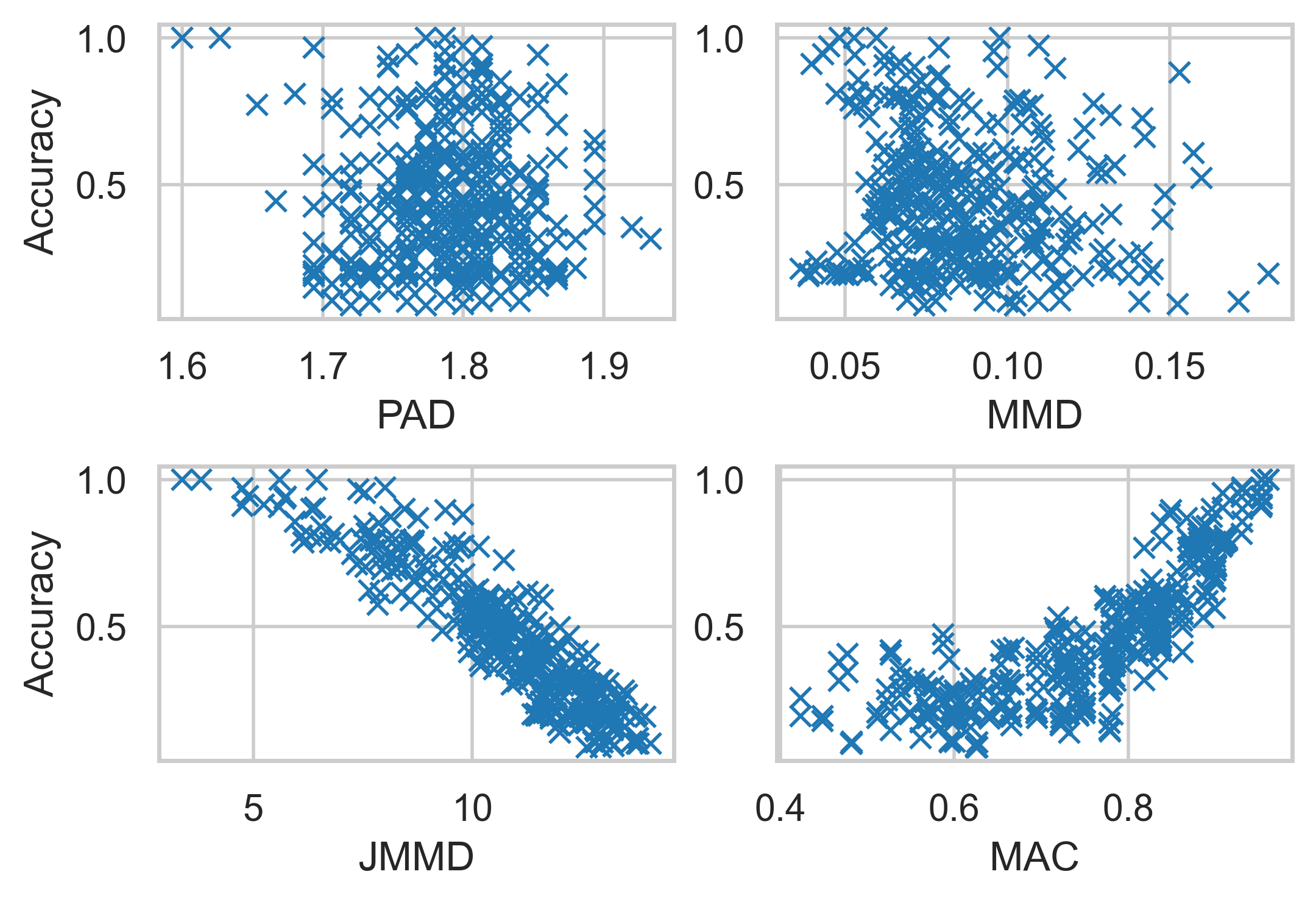}
\caption{A comparison of the accuracy of a damage classifier after NCA for each pair of structures in the numerical population, with two unsupervised data-based measures, the MMD and PAD, a supervised data-based measure, the JMMD, and the MAC-discrepancy. The PAD, MMD, and JMMD are zero only when the source and target distributions are the same, and the MAC will be one when all normal condition mode shapes are identical in both domains.}
\label{fig:metrix}
\end{figure}

A comparison of the similarity measures and accuracy is presented in Figure \ref{fig:metrix}. Although within the population there are a number of structures with sufficient similarity to achieve accurate predictions, neither of the unsupervised measures (the PAD and MMD) were indicative of the accuracy achieved via DA; this result is verified by their Pearson correlation coefficient with accuracy, given in Table \ref{tab:corr}. In addition, a number of tasks with low accuracy are associated with low values of the MMD, indicating that some cases follow similar marginal distributions but a number of clusters convey different contextual information (label switching); in some cases, this may have been caused by symmetry between the source and target. This result suggests that the population presented has several transfer tasks where conditional distribution similarity needs to be considered directly, and highlights the deficiency of the PAD and MMD in identifying related information in this setting. Moreover, this result illustrates a potential limitation with popular unsupervised TL algorithms, such as transfer component analysis (TCA) \cite{Pan2010} and the domain adversarial neural network (DANN) \cite{Ho2002}, as these methods rely on these measures for transfer.  In many real cases, structures will have differences in connections and local stiffness; thus, the fact that these changes may lead to conditional-distribution shifts presents a major challenge to knowledge sharing with unsupervised TL.\\

As expected, the supervised divergence measure, the JMMD, is well correlated with accuracy; however, this measure cannot be applied in practice since it requires label information for each damage-state in the target domain. However, the MAC-discrepancy also shows a strong correlation with accuracy\footnote{The correlation for the JMMD is negative since the lower value of the JMMD indicates more similar domains.}. In addition, the MAC-discrepancy and JMMD have a Pearson correlation coefficient of -0.84, suggesting that the MAC of the healthy structure could be used as a proxy for the JMMD to measure joint-distribution discrepancy in scenarios where labelled data are sparse or unavailable. Furthermore, whereas the JMMD requires labelled data for all classes in the target domain, the MAC-discrepancy only requires access to data from the healthy structure (as does NCA -- the DA method used here), meaning it could be estimated before any damage has been observed in the target structure.  \\

\begin{table}[h!]
    \centering
    \caption{The Pearson correlation coefficient between each of the similarity measures and the Accuracy (Acc), as well as the MAC and JMMD.}
    \begin{tabular}{lrrrrr}
    \toprule
    {} &    PAD-Acc &    MMD-Acc &   JMMD-Acc &    MAC-Acc  &    MAC-JMMD\\
    \midrule
    Correlation & 0.04 & -0.14 & -0.91 &  0.82  & -0.84\\
    \bottomrule
    \end{tabular}
    \label{tab:corr}
\end{table}

\section{Physics-informed feature selection for transfer learning}

Unsupervised TL is reliable only in scenarios where there is strong correspondence between the source and target conditional distributions, motivating the selection of features that meet this criterion. While the full set of features may not satisfy this condition, a subset might \cite{rojas2018invariant}. In general, the task feature selection involves finding a subset of features that can be used to effectively accomplish some downstream task (e.g. classification). In the case of unsupervised transfer learning, the goal of feature selection is to find a subset of features that satisfy the assumption that the features can be mapped into a shared feature space without using labelled target data. As discussed previously, supervised measures (e.g.\ the JMMD), are the gold-standard for indicating successful transfer. As it was also shown in Section 5 that the MAC is correlated with the JMMD, the current section introduces a feature selection based on the MAC to facilitate transfer in scenarios where domain adaptation assumptions hold only for a subset of features\footnote{The typical use of data-based measures, such as the MMD, would be to find a projection to a feature spaces where these measures are directly minimised \cite{JialinPan2011, Ganin2017}. However, the MAC cannot be used in this way, as it directly relates to the properties of a structure.
}.\\

\subsection{Physics-informed feature selection}

Standard approaches to feature selection typically incorporate a selection criterion and a search strategy to select a set of non-redundant features that maximise discriminative information; thus, reducing issues related to high feature dimension \cite{Murphy2014}. This paper introduces a transfer feature criterion (TFC), by incorporating the MAC-discrepancy into a feature-selection criterion to address the challenge of selecting a set of features that maximise the conditional distribution similarity between domains, such that conventional unsupervised DA methods can be applied reliably. In addition, balancing the trade-off between informative and domain-invariant features is a common challenge in TL \cite{Zhuang2021}; for this purpose, the source loss is included in the criterion; thus, the aim is to select a set of features to maximise the following objective function, 

\begin{equation}
    \mathcal{L} = -\frac{1}{n_s}\sum_{n=1}^{n_s} L(f_s(\mathbf{x}_{s,n}), y_{s,n}) + \lambda  d_{MAC}(\Phi_s,\Phi_t) - \mu C
\end{equation}

\noindent where $L(\cdot)$ represents the loss for a source predictive function $f_s(\cdot)$, $\lambda$ and $\mu$ are trade-off parameters, and $C$ represents a constraint to prevent the trivial solution of selecting the same feature multiple times; it is given by, \\
\begin{equation}
    C = \sum_{i=1}^d\sum_{i\neq j} [\mathbf{v}_{s,i} = \mathbf{v}_{s,j}] + [\mathbf{v}_{t,i} = \mathbf{v}_{t,j}] 
\end{equation}
where $[\cdot]$ represents the Iverson bracket, which takes the value $1$ if the values are equal, otherwise, it is 0. To ensure that the most similar source and target features are in correspondence, the target features are selected as, 
\begin{equation}
    \mathbf{v}_t = \underset{j}{\operatorname{argmax}} \ \text{MAC}(i,j)
\end{equation}

A search strategy is required to find a set of source indices. Feature selection is an NP-hard optimisation problem, meaning an exhaustive search would be required to guarantee a globally optimal solution. In this paper, only small feature sets are considered; thus, a exhaustive search is conducted. However, as the number of features increases, an exhaustive search will become computationally expensive, particularly for larger sets of features \cite{holland1992adaptation}. To extend the application of the TFC to high-dimensional feature spaces, heuristic search algorithms could be used \cite{FarrarC.R.CharlesR.2013Shm:}. \\

An initial demonstration of this feature selection approach between a plate generated using a 2D and 3D FE model is presented in Appendix D, showing that only corresponding modes can be used to share information between domains.

\subsection{Case study: Numerical Population}

In order to demonstrate the effectiveness of the TFC, the numerical population from Section 5.1 was evaluated by comparing its performance against naively applying DA to all the available features (natural frequencies). In addition, a multi-task approach is suggested to address issues with hyperparameter selection when target labels are unavailable.

\subsubsection{Multi-task learning for hyperparameter selection}

The lack of labels in unsupervised TL means that traditional hyperparameter selection schemes, such as cross-validation, are challenging to apply. To ensure that the TFC and the benchmark algorithms are appropriately tuned for a given target task, a multi-task approach was taken, performing joint empirical risk minimisation over a number of labelled source domains \cite{ando2005framework}. A subset of structures was assumed to be labelled, representing a number of source structures. By considering each pair of source structures as a source/ target pair, hyperparameter selection could be performed to find the best-performing model across all tasks. In this way, hyperparameters can be evaluated across $P$ tasks by,
\begin{equation}
    \mathbf{\theta} =  \underset{\mathbf{\theta} \in \Theta}{\arg\min} \sum^{P}_{p=1} \sum_{n=1}^{n_{t,p}} L(f_s(\mathbf{x}^p_{t,n}), y^p_{t,n})
\end{equation}

\noindent where $\mathbf{\theta}$ represents the vector of hyperparameters, while $\mathbf{x}^p_{t,n}$, $y^p_{t,n}$, and $n_{t,p}$ denote a feature vector, label, and the number of samples, respectively, for the target domain in task $p$. As discussed by Ando \emph{et al.}\cite{ando2005framework}, learning a set of hyperparameters on a single domain will likely overfit for the given task; however, utilising multiple related tasks allows for a general structure to be learnt, which can generalise to new tasks. In this paper, a number of structures are used to learn a general model that performs well across the population. A grid search was used for optimal parameter selection, but more efficient optimisation approaches could be used to reduce computation time.\\

\subsubsection{Transfer learning}

By applying the TFC, the objective is to find a set of features that satisfy the assumption that the conditional distributions are related, such that when it is used in conjunction with conventional unsupervised DA methods, a feature space can be found where the distributions are sufficiently close to transfer a classifier trained using only source labelled data. \\

Since it is robust to class imbalance issues \cite{poole2022statistic}, this paper uses NCA as the main method to perform unsupervised DA. To show the effectiveness of this approach in comparison to finding more complex (nonlinear) mappings via unsupervised DA,  two DA algorithms that perform dimensionality reduction and have been previously used in SHM were selected to benchmark the feature-selection approach -- TCA and \emph{balanced distribution adaptation} (BDA); they are briefly summarised here:

\begin{itemize}[leftmargin=1.5em, itemsep=0em]
    \item TCA \cite{JialinPan2011}: learns a feature space that minimises the MMD \cite{Gretton2012a}, by MMD-regularised principal component analysis (PCA). TCA is capable of learning flexible nonlinear mappings; therefore, it was used to verify that in the presented case studies, unsupervised DA is limited since minimising marginal-distribution shift alone may be insufficient.
    \item BDA \cite{Wang2017}: extends TCA by also attempting to minimise the MMD between the class-conditional distributions $p_s(\mathbf{x}|y)$ and $p_t(\mathbf{x}|\hat{y})$, where $\hat{y}$ are pseudo-label predictions. Since there are no labels in the target, BDA uses pseudo labels from predictions to assign target samples to a given class and finds the distance between the class-conditional distributions. As such, BDA is used as a benchmark to show that unsupervised methods that attempt to minimise the conditional distribution discrepancy are unable to provide significant performance increases if the conditional distributions are not sufficiently related in the original feature space.
\end{itemize}

Following recent studies suggesting that NCA is an important preprocessing step for TL in SHM, it was applied prior to the application of each algorithm \cite{poole2022statistic, giglioni2024domain, wickramarachchi2024damage, giglioni2025transfer, delo5006357influence}; for conciseness the importance of statistic alignment is not further discussed in this paper and the interested reader may refer to \cite{poole2022statistic} for a more in-depth discussion. These algorithms were also applied to the TFC-selected features, following NCA, to assess the potential benefit of further reducing distribution shift via more flexible mappings. \\

Hyperparameter selection for all models was performed using the multi-task approach, with five structures being utilised; the associated tasks were considered ``validation tasks" and were not included in the final results, reducing the total tasks for testing from 360 to 342. \\

Each model has a number of parameters that may influence the results; these  selected were via the multi-task cross validation scheme. For the TFC, this is the number of selected features and trade-off parameters for the MAC in the loss function (Equation 5); these were selected as $D=7$ and $\lambda=0.1$. The kernel-based algorithms also require a regularisation term and the number of latent features to be specified; these were chosen as $D=9$ and $\lambda=0.1$ for TCA and $D=5$ and $\lambda=0.1$ for BDA. When applied after the TFC, TCA and BDA were chosen to reduce the feature dimension by one. Additionally, BDA has a trade-off parameter between the marginal and class-conditional distributions, but since the objective of implementing this algorithm was to investigate the use of pseudo-label-based class-conditional MMD, this was set to unity, so the marginal MMD was not used. Finally, as these methods utilise an RBF kernel, a length scale must be specified; to reduce the hyperparameter search space the median heuristic was used \cite{Gretton2012a}.  Classification performance was evaluated using accuracy scores and the JMMD was used to quantify distribution discrepancy, each were calculated using the test data. It should be noted that feature dimension may also influence the JMMD. \\

\subsection{Results: unsupervised transfer learning}

The mean accuracy on the target for all transfer tasks is presented in Table \ref{tab:mean_change}, including a comparison with performing no DA. By only applying a linear  transform estimated by NCA it can be seen that classification accuracy improved significantly. Whilst the accuracy after applying NCA is still relatively low (45\%), it is important to emphasise that this case study includes transfer tasks across a range of structures; thus, many transfer tasks are challenging as some pairs of structures have very different responses to damage. It can be seen that even with these challenging transfer tasks that NCA results in an improvement across the entire population with a low rate of negative transfer. Furthermore, when NCA and the TFC are applied together, further improvements to the average classification accuracy can be seen.  \\

Here both TCA and BDA do not provide a significant improvement compared to only applying the linear transform found via NCA, suggesting neither a more complex nonlinear mapping found via the MMD (TCA) or the class-conditional MMD using pseudo-labels (BDA), can reliably provide further improvement to generalisation when some features have a different response to damage. This result is perhaps because these methods cannot find a shared space where conditional distribution distance is low by only relying on unlabelled data to learn a mapping. On the other hand, the TFC (which was applied in conjunction with NCA), was able to significantly improve classification results on average, indicating that by selecting features with a similar response to damage via the MAC (similar conditional distributions), conventional approaches to DA (NCA) can more effectively facilitate transfer. Since conditional-distribution shift cannot typically be reduced using data-based methods without target labels, this highlights a potential benefit of leveraging physics-knowledge.  \\

\begin{table}[t!]
\caption{Mean accuracy and JMMD for all tasks for the source and target test data on the numerical population of structures.}
\label{tab:mean_change}
\centering
\begin{tabular}{lrrrrrrr}
\toprule
{} &  no DA &    NCA &   TCA &   BDA &    TFC &  TFC+TCA &  TFC+BDA \\
\midrule
source test &   1.00 &   1.00 &  1.00 &  0.99 &  1.00 &    1.00 &    0.99 \\
target test &   0.10 &   0.45 &  0.45 &  0.45 &  0.58 &    0.56 &    0.61 \\
JMMD        &  19.01 &  10.77 &  9.92 &  7.17 &  8.76 &    7.90 &    5.04 \\
\bottomrule
\end{tabular}
\end{table}

\begin{figure}[b!]
    \centering
    \begin{subfigure}[b]{0.46\textwidth}
         \centering
        \includegraphics[width=\textwidth]{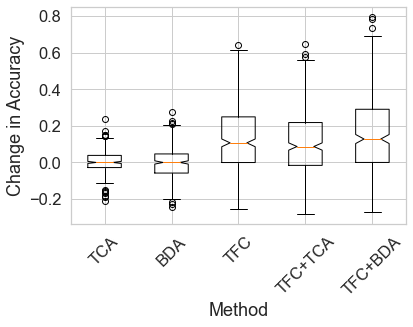}
         \caption{}
         \label{fig:hetro}
     \end{subfigure}
    \begin{subfigure}[b]{0.45\textwidth}
         \centering
         \includegraphics[width=\textwidth]{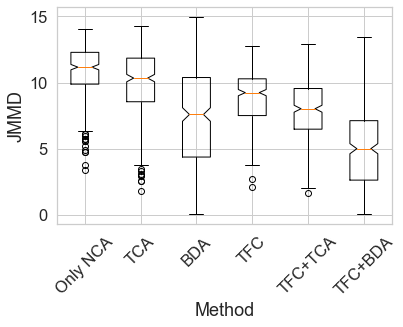}
         \caption{}
         \label{fig:fsd}
     \end{subfigure}
    \caption{Box-and-whisker plots showing the change in accuracy compared to NCA (a) and the JMMD after each transfer method (b) for the numerical case study.}
    \label{fig:boxxy}
\end{figure}

After finding a more similar set of features via the TFC, BDA was able to further improve the average accuracy of classification. This improvement may be because BDA relies on accurate estimation of pseudo-labels, which becomes more feasible after discarding less related features.\\

Figure \ref{fig:boxxy}(a) presents a box-and-whisker diagram showing the range of change in accuracy of target classification compared to only applying NCA. It can be seen that for both TCA and BDA, there is a relatively high rate of negative transfer, meaning accuracy across the population is not improved. Conversely, the TFC lead to a low rate of negative transfer, improving generalisation for a much larger portion of the population, as well as providing larger maximum improvements to classification accuracy.  \\

Another consideration is that both TCA and BDA use (unlabelled) feature data for each damage class; this may not be a realistic assumption in engineering scenarios, where often only limited damage data are available in the target. This paradigm is called partial DA, where conventional DA is known to be prone to negative transfer \cite{Cao2018a}. The TFC only relies on using the mode shapes from normal condition, as does the DA method used here (NCA); consequently, it could be applied to practical scenarios to facilitate real-time health-state prediction in structures with no previously observed damage. \\

 The JMMD after applying each method is presented in Figure \ref{fig:boxxy}(b) to show the discrepancy between the joint distributions after alignment.  TCA and BDA both reduce distribution distance (Figure \ref{fig:boxxy}(b)). However, these methods do not improve classification results across the population, as classes may be misaligned if the conditional distributions are unrelated, but marginal-distribution discrepancy will still be reduced. After applying the TFC, the JMMD still indicates a discrepancy between the data \cite{wang2019characterizing}. Applying the TFC and the DA algorithms concurrently further reduces discrepancy, particularly with BDA, which could lead to better generalisation.  \\

\begin{figure}[b!]
    \centering
         \centering
        \includegraphics[width=0.5\textwidth]{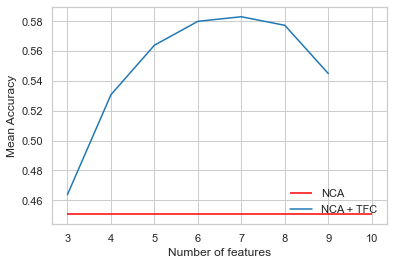}
    \caption{Mean accuracy for transfer when using the TFC to select a varying number of features on the numerical population. The accuracy of NCA using all the available features (ten natural frequencies) is shown in red and the accuracy for the TFC selecting a subset of features is shown in blue.}
    \label{fig:feat_sens}
\end{figure}

To investigate the impact of the number of selected features and the sensitivity of the hyperparameter selection scheme, the mean accuracy was evaluated when varying numbers of selected features; Figure \ref{fig:feat_sens} presents these results. Overall, discarding dissimilar features is largely beneficial, improving performance across the population. It can also be seen there is a balance between discarding dissimilar features, and retaining informative features (discriminative information), where in this case study an optimal average accuracy across population is given by seven features. Using a small subset of five structures, the multi-task hyperparameter selection scheme was able to select this optimal number of features.   
\\ 

\section{Experimental case study: heterogeneous helicopter blades}

To further explore the application of using modal similarity to inform transfer, a case study consisting of two heterogeneous full-scale helicopter blades is presented. Specifically, the blades are from a Robinson R44 and a Gazelle helicopter. These blades were chosen using engineering judgement, as both blades are similar in size and internal structure, suggesting there is potential to share information\footnote{As previously mentioned, in practice it would be desirable to use a principled approach to select source/target domains, particularly for complex structure. However, to identify two similar, but distinct blades, engineering judgement was sufficient to demonstrate the benefit of the TFC in this case. More principled domain selection approaches are left for future work. }. Importantly, there are several discrepancies, motivating the application of TL; these differences are summarised in Table \ref{tab:blades}. \\

\begin{table}[t!]
\centering
\caption{Summary of the key differences between the Robinson R44 and Gazelle blades.}
\label{tab:blades}
\begin{tabular}{lllllll}
\cline{1-7}
                                   &  Material & Mass & Length  & Width  & Leading edge & Trailing edge  
                                      \\&& (kg)&(m)&(m)&thickness (mm)&thickness (mm)
                                      \\ \hline
\multicolumn{1}{l}{Metal blade}     & steel     & 26.95        & 4.88       & 0.255     &    32.70 & 4.30          \\ 
\multicolumn{1}{l}{Composite blade} & carbon fibre  & 37.00        & 4.83       & 0.300       &    28.10 & 1.00               \\ \hline
\end{tabular}
\end{table}

\begin{figure}[b!]
    \centering
    \includegraphics[width=0.6\textwidth]{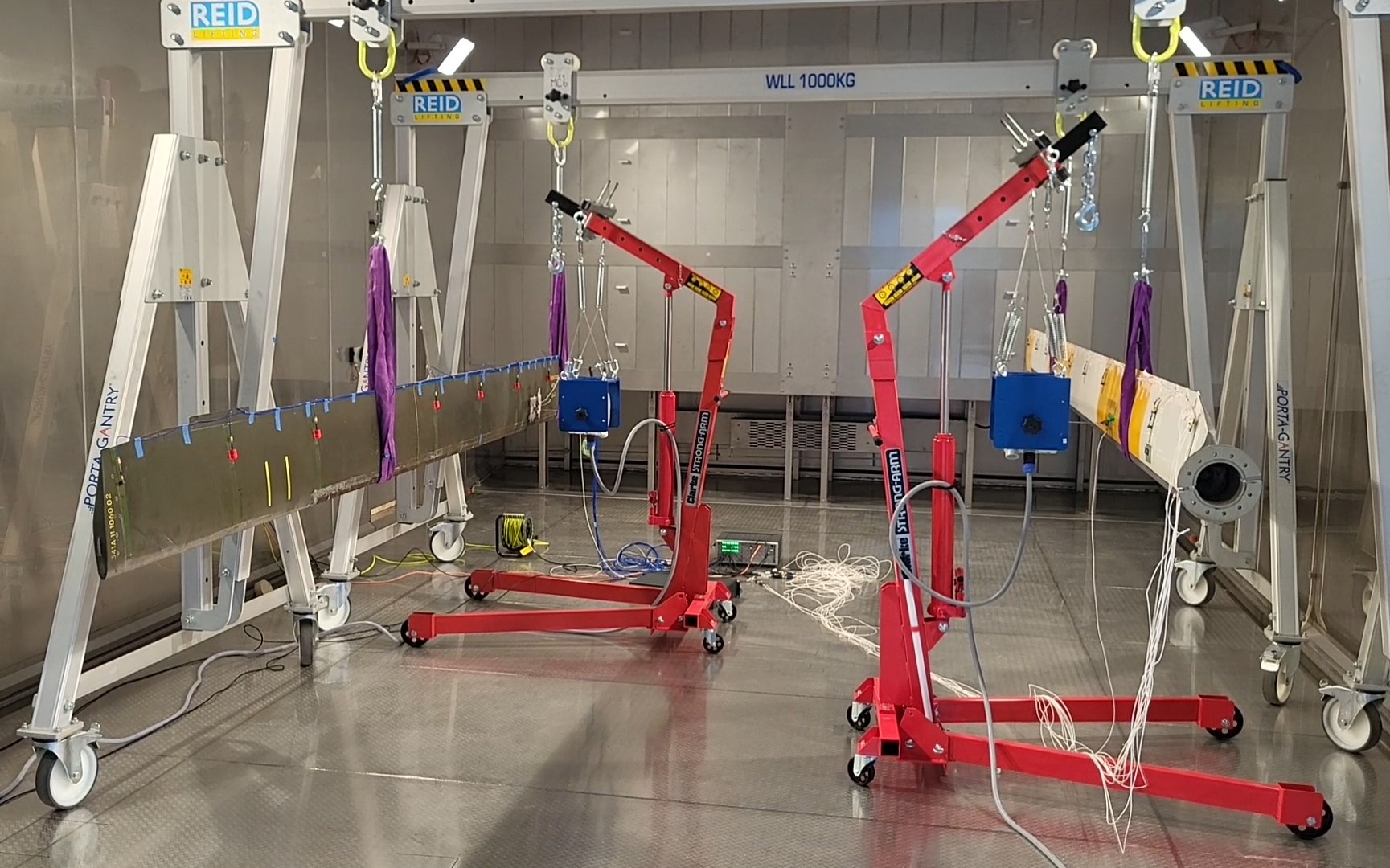}
    \caption{The experimental setup to perform modal testing on a metal (right) and composite (left) blade simultaneously.}
    \label{fig:piccy}
\end{figure}

The experimental set-up is shown in Figure \ref{fig:piccy}. Modal testing was conducted on the blades in a free-free configuration, utilising electrodynamic shakers attached in the flapwise direction, applying a continuous pink noise random excitation up to 800Hz, with a decay rate of 3dB/Octave and a sample rate of 1600Hz. Data were collected via ten uniaxial 100 mV/g accelerometers,  placed on the underside of each blade along the length, at positions corresponding to the same non-dimensionalised length and width, the location of the sensors and shakers are given in Appendix E. To mitigate noise effects, ten frequency-domain averages were obtained. Testing was conducted on both blades simultaneously, assuring that data from both blades corresponded to the same environmental conditions.\\

Data were collected for five health-states, including the normal condition and four pseudo-damage states,  relating to adding small masses to specific locations along the length. The added masses were positioned at standardised lengths and widths of the blades and the size of the mass was scaled to maintain a consistent ratio between the added mass and blade mass. As such, the locations of damage should correspond to similar points of a given mode shape and the extent of ``damage" can be considered equivalent for both blades. A summary of the datasets is given in Table \ref{tab:set}, where L* and W* refer to the non-dimensionalised length and width respectively, which were measured from the root and leading edge.  A more detailed outline of the test regime is presented in Appendix E. It should be noted that the relationship between damage sensitivity and the position of the nodes (as discussed in Section 4) should be inverted for added masses, with the masses having the largest influence on the modes when placed close to anti-nodes.\\

Added masses were used to represent damage here as they provide a controlled, repeatable way of “damaging” the structure. These health-state allow for the proposed methodology to be validated for transferring a classifier predicting the location of a mass increase, which is equivalent to a reduction in stiffness in linear systems. However, one limitation is that various damage modes structures regularly experience may introduce more complex nonlinear behaviour; the investigation of the use of the proposed approach is left to future work.  Note that the relationship between damage sensitivity and the position of the nodes (as discussed in Section 4) should be inverted for added masses, with the masses having the largest influence on the modes when placed close to anti-nodes. It is important to note that, while added masses are expected to result in similar change in response in these blades, many common damage modes would exhibit either very different effects in each blade, or may not be relevant to both blades; for example, crack propagation in metals and composites will not have similar behaviours. Further investigation into specifically which damage modes have similar enough behaviour between structures for the purpose of transfer learning is left to future work. \\

\begin{table}[t!]
\centering
\caption{Summary of the blade datasets. The mass ratio between the metal and composite blade is 0.728.}
\label{tab:set}
\begin{tabular}{llllll}
\toprule
Mass  & Repeats & Mass  & Metal & Comp  & Mass \\ state &  & location (L*, W*) & mass (g) & mass (g) & ratio \\
\midrule
Normal condition         & 25      & -                      & -              & -             & -          \\
Damage 1         & 10      & (0.627, 0.577)         & 76.6           & 105.8         & 0.724      \\
Damage 2         & 10      & (0.876, 0.577)         & 76.6           & 105.8         & 0.724      \\
Damage 3        & 10      & (0.627, 0.577)         & 250.0            & 350.0           & 0.714      \\
Damage 4         & 10      & (0.876, 0.577)          & 250.0            & 350.0           & 0.714      \\
\bottomrule
\end{tabular}
\end{table}

\subsection{Transfer  learning methodology}

In this case study, the objective was to classify the normal and four damaged states of the blades, with the aim of transferring the acquired knowledge from one blade to another. Two tasks were considered, wherein each blade was considered as both the source and the target domain. These tasks will be referred to as M$\rightarrow$C when considering the metal blade  as the source and composite blade the target, and C$\rightarrow$M for the opposite case.\\

\begin{figure}[t!]
    \centering
    \begin{subfigure}[b]{0.6\textwidth}
         \centering
        \includegraphics[width=\textwidth]{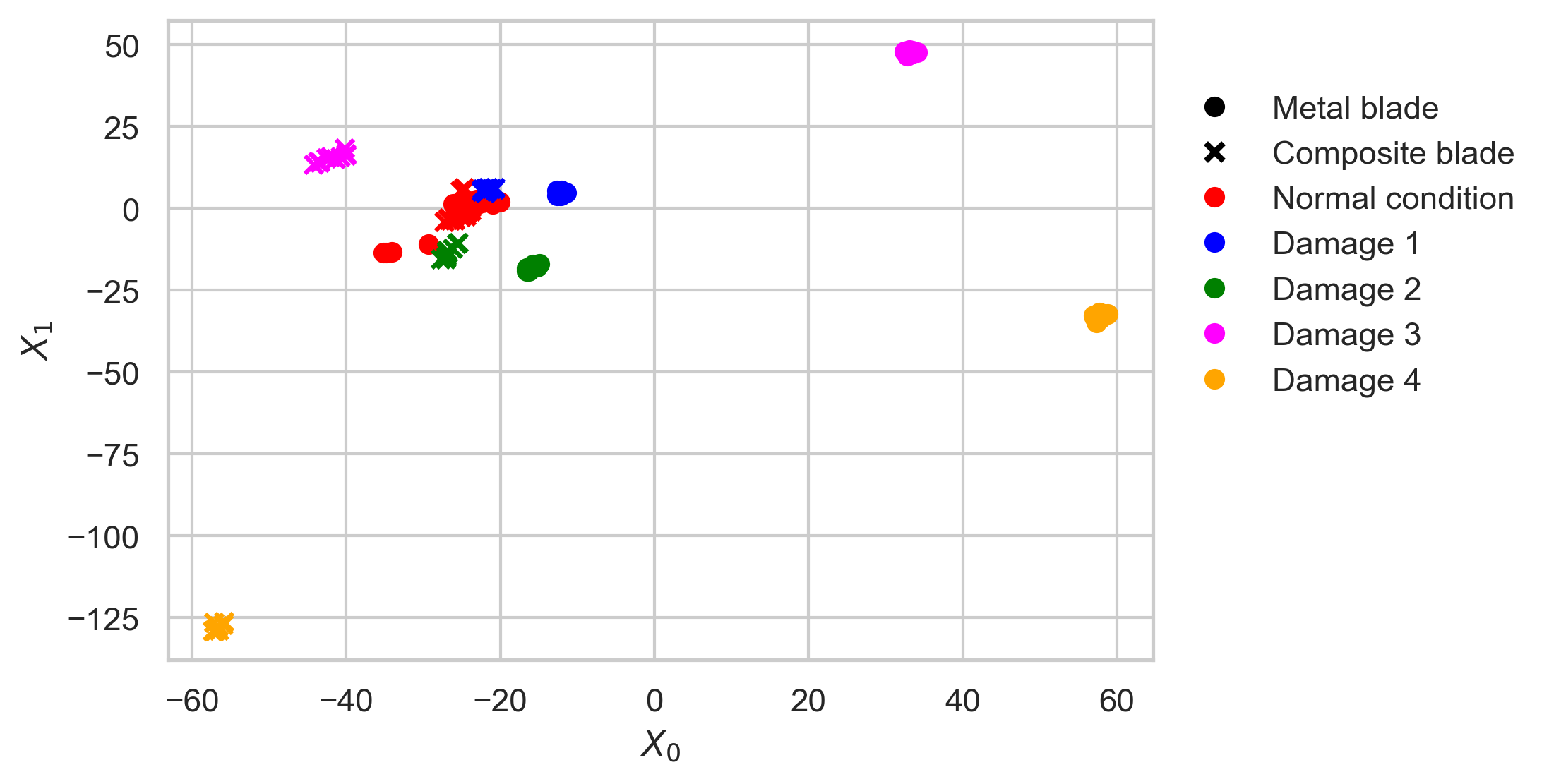}
         \caption{}
         \label{fig:hetro}
     \end{subfigure}
    \begin{subfigure}[b]{0.6\textwidth}
         \centering
         \includegraphics[width=\textwidth]{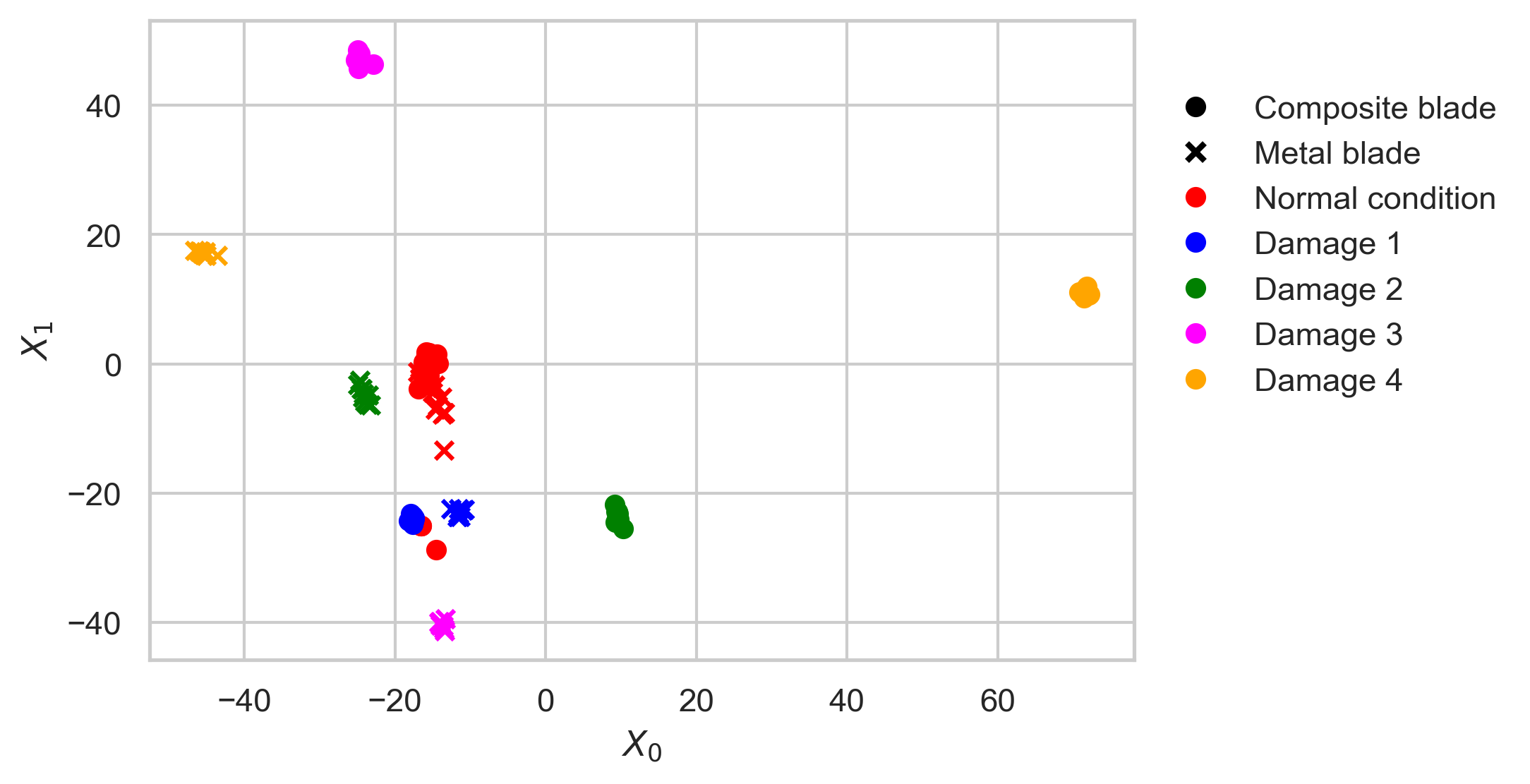}
         \caption{}
     \end{subfigure}
    \caption{PCA of all FRF features up to 250Hz, learning the principal components on the metal blade data (a) and composite blade data (b). The source and target data are represented by (\protect$\medcircle$) and (\protect$\times$) respectively. Normal condition is shown by red markers, while the pseudo-damage states induced with the 76.6g mass are indicated by blue and green markers, and the pseudo-damage induced with the 250g masses are shown in magenta and yellow.}
    \label{fig:hetro_pca}
\end{figure}

An example of a frequency response functions (FRFs) of each blade for the normal condition, taken from the sensor closest to the tip (sensor 10, see Appendix E) is presented in Figure \ref{fig:FRF_comparison}. Initially, it can be seen that there are significant differences between the responses of the blades; the peaks are shifted and their amplitudes vary. To visualise the datasets, NCA was performed on the raw FRF data to correct for differences in mean and scale of the FRF amplitude. Following adaptation via NCA, PCA was learnt on a single domain and was applied to both datasets; the first two principal components are presented for PCA learnt using data from the metal blade in Figure \ref{fig:hetro_pca}a and on the composite blade in Figure \ref{fig:hetro_pca}b. It can be seen in both cases that the average distance between the normal condition and damage classes is generally larger in the source, suggesting PCA trained on one set of FRF features does not effectively capture the variance in the other. This discrepancy may largely be influenced by features not being in correspondence i.e. a frequency with a large contribution from a given mode in one blade is often in direct comparison with a frequency that has little contribution from any modes in the other blade, meaning they will have different sensitivities to damage. Thus, the frequencies with high damage-sensitivity (discriminative features) differ when the FRF frequencies are not processed to account for the difference in natural frequencies.\\

\begin{figure}[h!]
    \centering
    \includegraphics[width=0.6\textwidth]{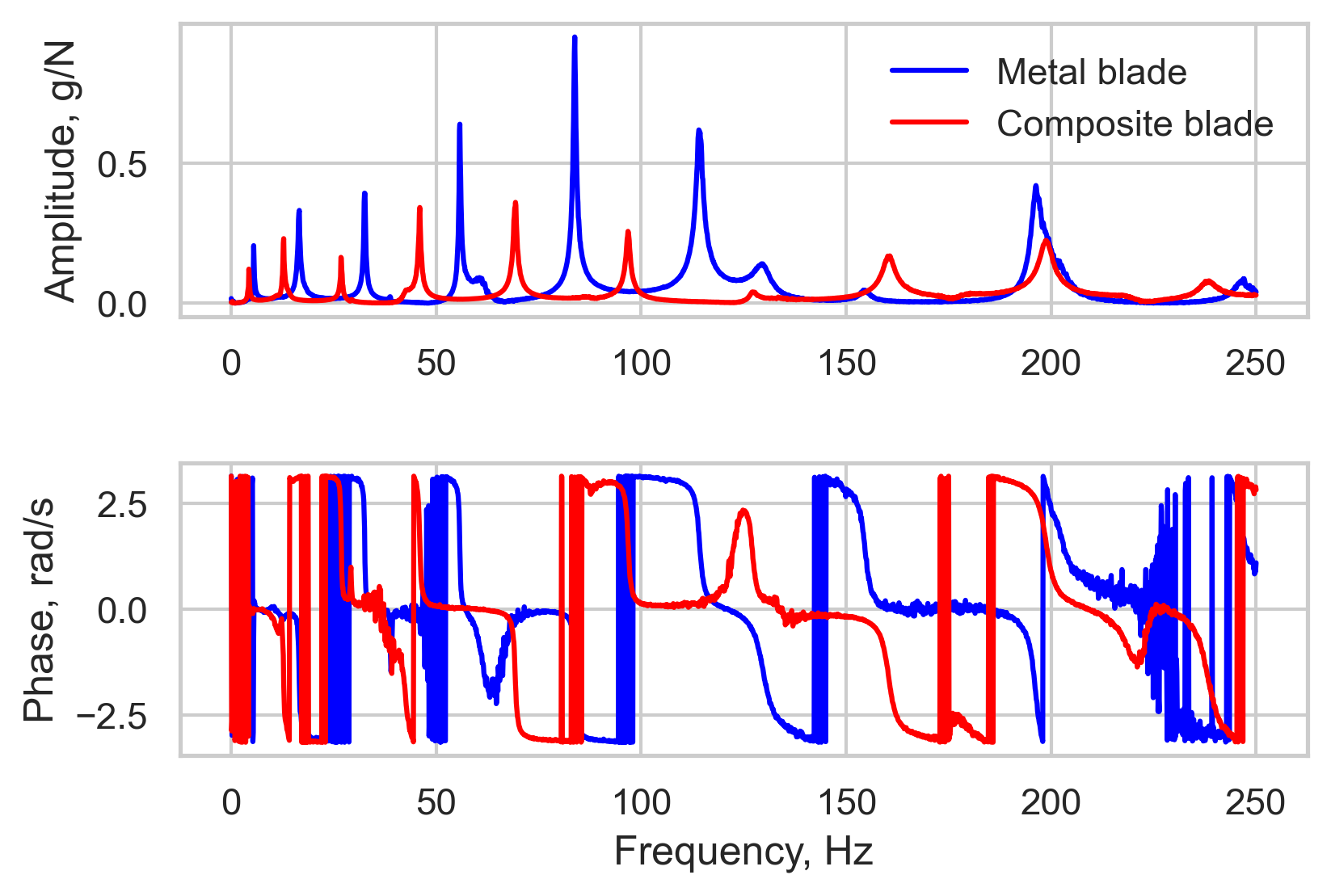}
    \caption{An example of the FRFs for the metal blade (blue) and composite blade (red) with no added masses.}
    \label{fig:FRF_comparison}
\end{figure}

\begin{figure}[t!]
    \centering
\includegraphics[width=0.55\textwidth]{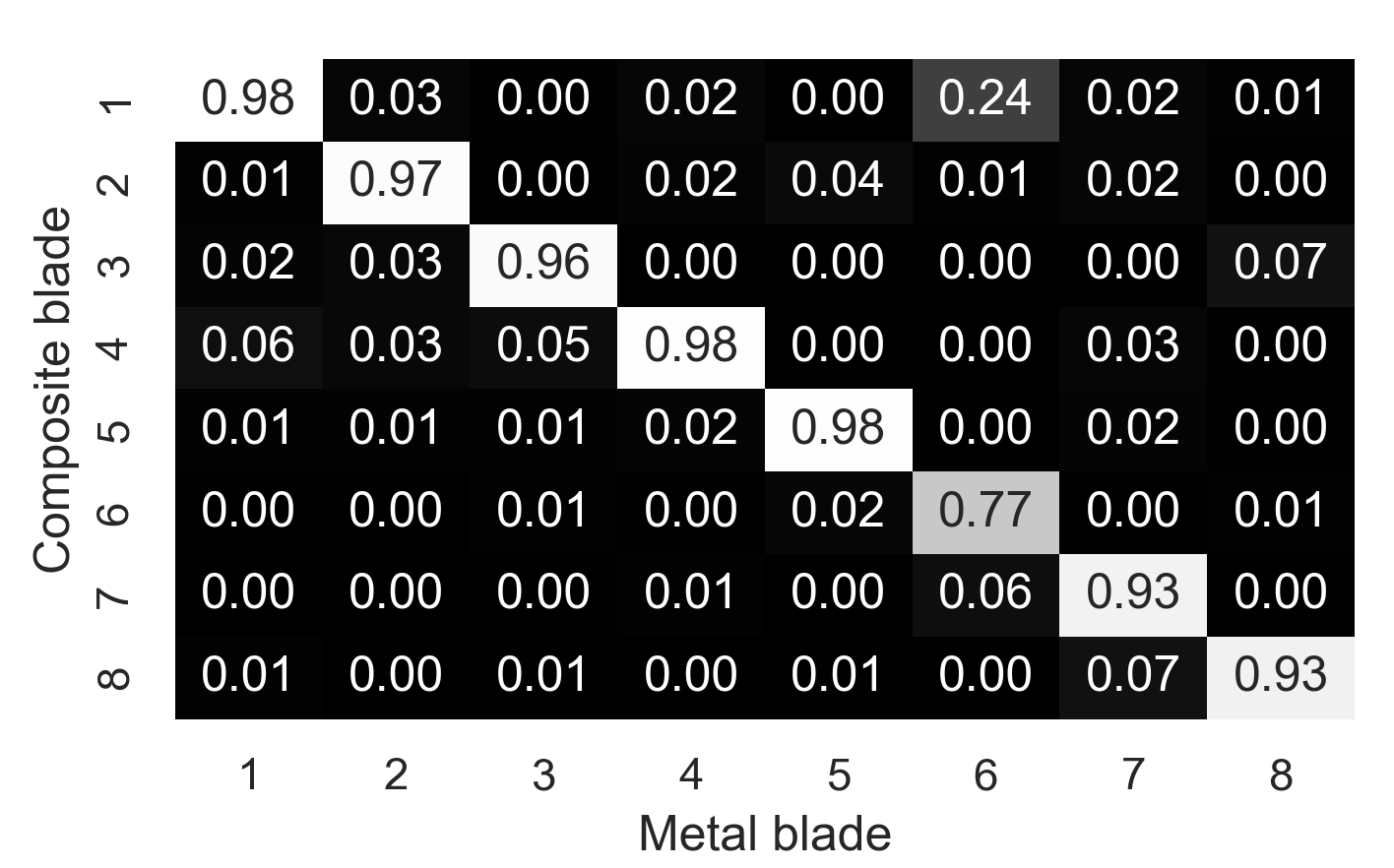}
    \caption{MAC matrix between the modes of the metal blade and composite blade normal condition.}
    \label{fig:machet}
\end{figure}

The FRF amplitude data from the sensor closest to the tip were utilised as features to expedite the modal analysis process. A subset of the FRF was determined by selecting a window of 20 features centred around the natural frequencies; thus, in using these features, peaks of the FRF are compared across domains, which means features correspond to a similar physical phenomenon. As such, modal analysis was only conducted three times on the normal-condition data, in order to identify the regions of the FRF that should be put in correspondence for effective transfer, which was assumed to predominantly correspond to the respective mode; an example of this feature space is shown in Figure \ref{fig:FRF_comparison2} and the natural frequencies are given in Appendix E, along with natural frequencies for the damage-states, which were not used in this analysis. In addition, the mode shapes were used to calculate the MAC between the blades, which were then used to inform feature selection; the MAC matrix is given in Figure \ref{fig:machet}.  Nine modes were identified in this range in the composite blade. However, since one of the benchmarks in the following section aims to test the efficacy of the TFC by including benchmark results for algorithms that use all potential features, the ninth mode identified in the composite blade was removed to maintain a homogeneous feature space between domains. Note that the first eight modes were already in correspondence, although this may not always be the case and in these scenarios using the MAC to bring modes into correspondence may be even more critical.\\

\begin{figure}[t!]
    \centering
    \includegraphics[width=0.8\textwidth]{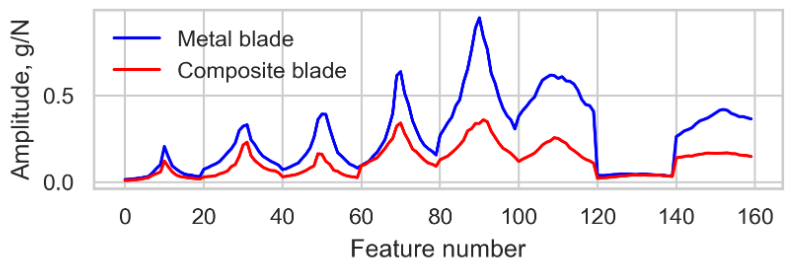}
    \caption{Example of the feature space after selecting a window of 20 frequencies centred around the natural frequencies for the metal blade (blue) and composite blade (red).} 
    \label{fig:FRF_comparison2}
\end{figure}

The transfer-learning methodology closely adhered to the one outlined in Section 6, with the inclusion of PCA, trained on the source domain, as a benchmark for a typical feature-reduction technique; feature dimension was selected to maintain 90\% of the variance. The primary deviation in the approach was in the hyperparameter selection and testing scheme. As the population lacked multiple tasks that could facilitate multi-task hyperparameter selection, the regularisation hyperparameters were selected as the values found in the numerical case ($\lambda=0.1$ for all cases). The number of features for each method was determined via leave-one-out (LOO) validation using the source data, under the assumption that if the modes are sufficiently discriminative in the source, they should also be in the target. This approach selects two modes for all methods. Since the data were limited, the test results were determined using LOO validation, excluding the JMMD values, which were found on all the data since this measure requires a distribution of data. It should be noted that the limited sample size of this case study may impact the reliability of the JMMD. The TL methods are benchmarked against a KNN trained without TL, where ``no DA" the KNN learnt using features as presented in Figure \ref{fig:FRF_comparison2}.

\subsection{Results}

\begin{figure}[b!]
    \centering
    \begin{subfigure}[b]{0.6\textwidth}
         \centering
        \includegraphics[width=\textwidth]{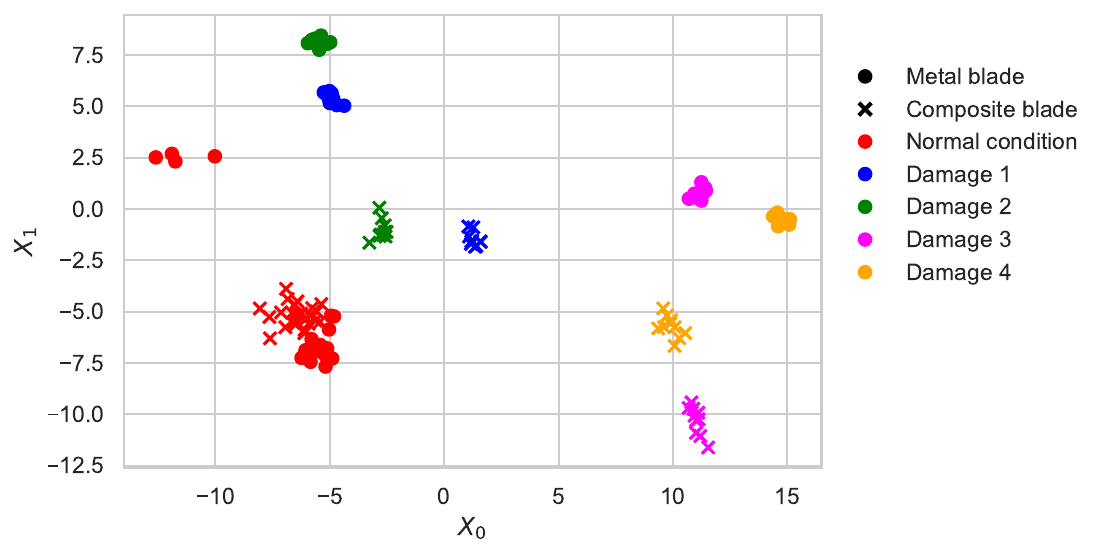}
         \caption{}
         \label{fig:hetro}
     \end{subfigure}
    \begin{subfigure}[b]{0.6\textwidth}
         \centering
         \includegraphics[width=\textwidth]{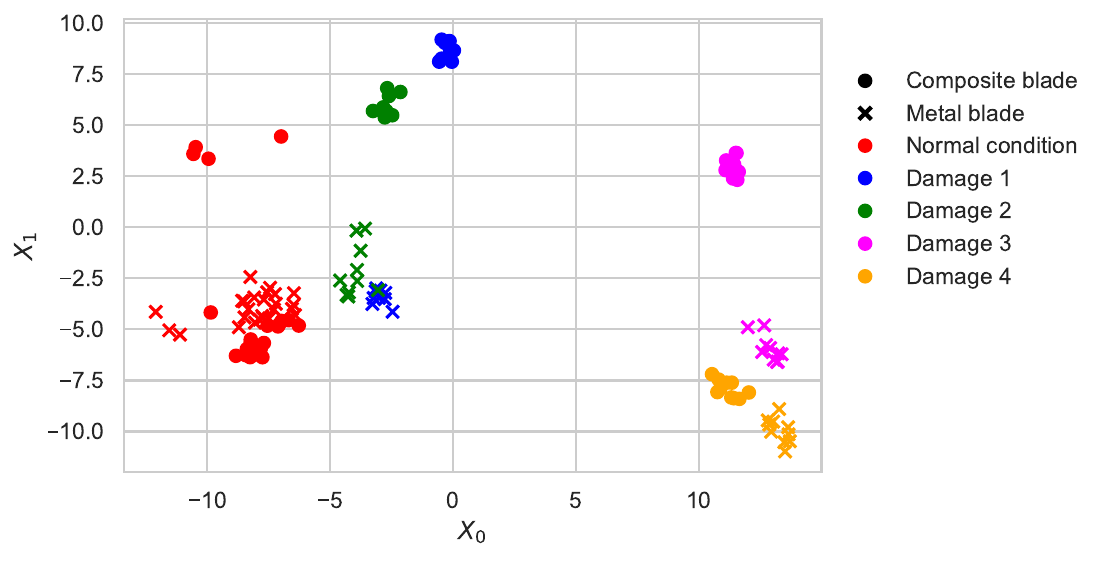}
         \caption{}
         \label{fig:hetro}
     \end{subfigure}
    \caption{PCA visualisation of all of the features selected in a window centred around the natural frequencies. (a) is for M$\rightarrow$C. The source data are shown by (\protect$\medcircle$) markers and the target data are shown by (\protect$\times$ markers), respectively.  Both the training and testing data are plotted for a one iteration of LOO validation.}
    \label{fig:PCA_align}
\end{figure}

The initial features are visualised using PCA in Figure \ref{fig:PCA_align}. It can be seen that each damage class (shown by their respective colour) is not necessarily close between domains, indicating differences in the data distributions, meaning it is unlikely a classifier would generalise well. This result may be because this initial feature space includes information from a less-similar mode (Mode 6) and a number of modes that are likely not damage-sensitive (e.g. the lower modes). It can also be seen that four normal-condition data are shifted in both domains. These data relate to the last four measurements, which were taken after the majority of tests (refer to Appendix E for more details). As such, these differences may be caused by changes in the boundary conditions, from interacting with the blades during testing or variations in temperature. The higher scatter in the normal-condition data means that it is pertinent that features are both discriminative and have high cross-domain similarity. \\

 \begin{table}[t!]
      \centering
 \caption{Mean accuracy for the source and target test data and the JMMD between all data for transfer between the metal and composite blade datasets. }
\label{tab:het_acc}
\begin{tabular}{lrrrrrrrrr}

\toprule
{} &  No DA &   NCA &   PCA &   TCA &   BDA &    TFC &  TFC+ &  TFC+\\
{} &   &    &    &    &    &     &  TCA &  BDA\\
\midrule
M$\rightarrow$C: Source Test Accuracy &   1.00 &  0.98 &  0.98 &  1.00 &  1.00 &  1.00 &  0.98 &  1.00  \\
M$\rightarrow$C: Target Test Accuracy &   0.15 &  0.71 &  0.69 &  0.54 &  0.75 &  0.85 &  0.88 &  1.00 \\
M$\rightarrow$C: JMMD            &   9.66 &  5.50 &  5.90 &  5.04 &  0.73 &  3.58 &  0.74 &  0.27 \\
C$\rightarrow$M: Source Test Accuracy &   1.00 &  1.00 &  1.00 &  1.00 &  1.00 &  1.00 &  1.00 &  1.00  \\
C$\rightarrow$M: Target Test Accuracy &   0.38 &  0.58 &  0.58 &  0.77 &  0.85 &  0.89 &  0.85 &  1.00 \\
C$\rightarrow$M JMMD       &   9.66 &  5.42 &  4.24 &  3.76 &  1.73 &  3.85 &  1.59  &  0.26  \\
\bottomrule
\end{tabular}

 \end{table}
 
Table \ref{tab:het_acc} shows the test accuracy on the target obtained from LOO validation and the JMMD, which was obtained using all the source and target data. It can be observed that in comparison to naively trying to apply a classifier trained using the source labels (no DA), applying DA via NCA led to an improvement in classification. While all test samples were correctly classified in the source domains for each task, the accuracy in the target domain for no DA and NCA was significantly compromised. This result suggests that transferring a classifier from the source domain generalises worse than supervised learning using target data,  which is indicative of domain shift leading to large generalisation errors. Furthermore, the results achieved after also applying PCA with NCA (PCA in Table \ref{tab:het_acc}), suggest that this issue cannot be alleviated by reducing the feature dimension alone. In addition, further unsupervised DA after applying NCA (TCA and BDA), did further improve classification in the target in some cases, although TCA lead to negative transfer in M$\rightarrow$C.\\

\begin{figure}[b!]
    \centering
    \begin{subfigure}[b]{0.6\textwidth}
         \centering
        \includegraphics[width=\textwidth]{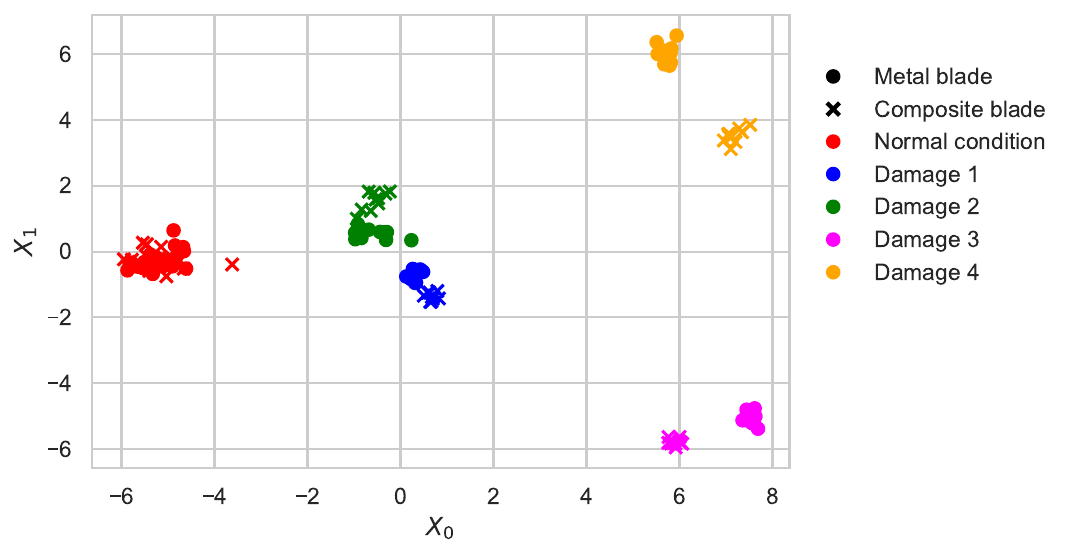}
         \caption{}
         \label{fig:hetro}
     \end{subfigure}
    \begin{subfigure}[b]{0.6\textwidth}
         \centering
         \includegraphics[width=\textwidth]{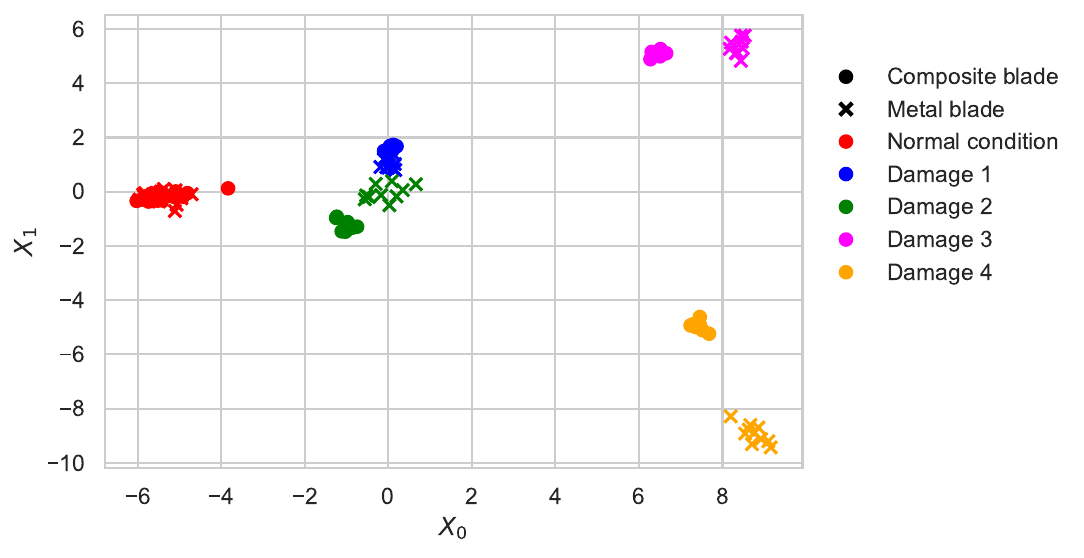}
         \caption{}
         \label{fig:hetro}
     \end{subfigure}
    \caption{PCA visualisation of the TFC-selected frequencies, corresponding to the fourth and fifth modes, for M$\rightarrow$C (a) and C$\rightarrow$M (b). The source data are shown by (\protect$\medcircle$) markers and the target data are shown by (\protect$\times$ markers), respectively.  Both the training and testing data are plotted for a one iteration of LOO validation.  }
    \label{fig:PCA_select}
\end{figure}

\begin{figure}[b!]
    \centering
    \begin{subfigure}[b]{0.36\textwidth}
         \centering
        \includegraphics[width=\textwidth]{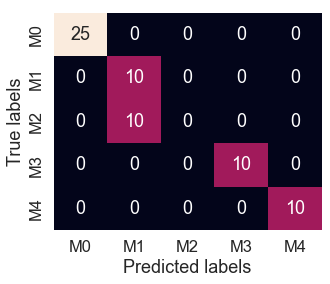}
         \caption{}
         \label{fig:hetro}
     \end{subfigure}
    \begin{subfigure}[b]{0.36\textwidth}
         \centering
         \includegraphics[width=\textwidth]{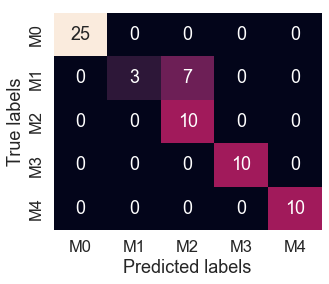}
         \caption{}
         \label{fig:hetro}
     \end{subfigure}
    \caption{Confusion matrices for the test data on the TFC selected features using a KNN for M$\rightarrow$C (a) and C$\rightarrow$M (b). }
    \label{fig:confuseee}
\end{figure}

\begin{figure}[t!]
    \centering
    \begin{subfigure}[b]{0.6\textwidth}
         \centering
        \includegraphics[width=\textwidth]{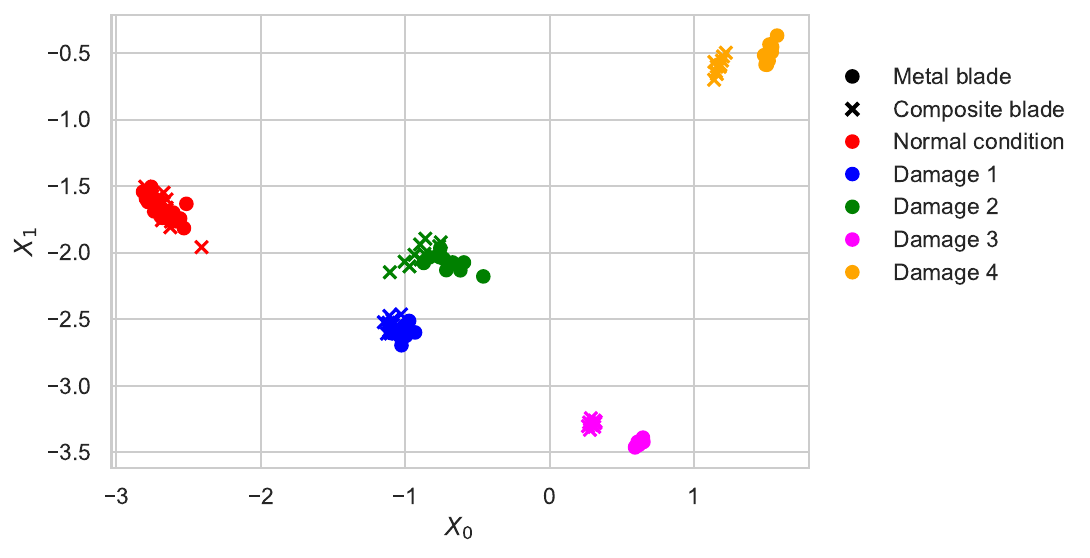}
         \caption{}
         \label{fig:hetro}
     \end{subfigure}
    \begin{subfigure}[b]{0.6\textwidth}
         \centering
         \includegraphics[width=\textwidth]{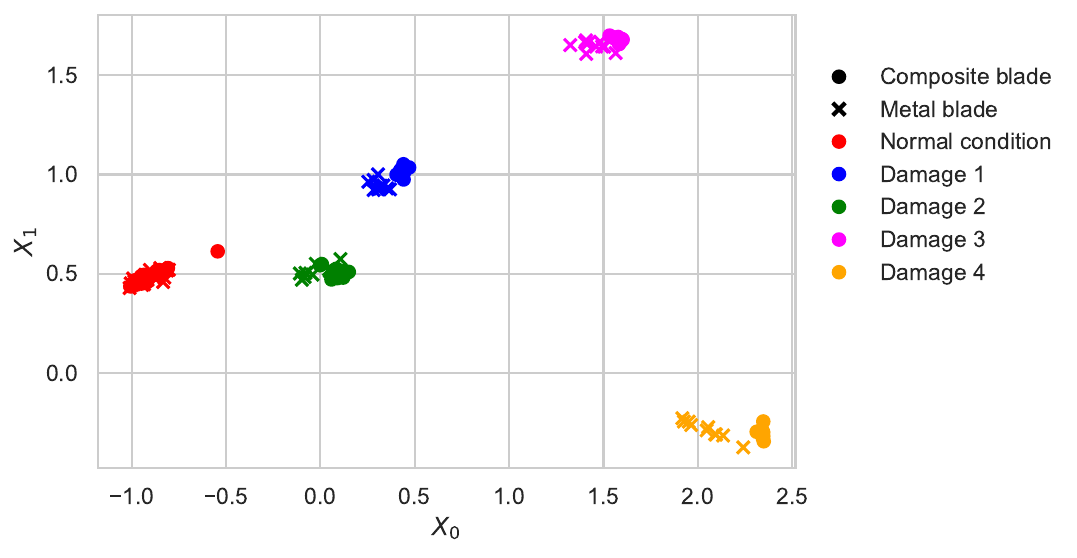}
         \caption{}
         \label{fig:hetro}
     \end{subfigure}
    \caption{Features found via BDA applied to the TFC-selected frequencies, corresponding to the fourth and fifth modes, for M$\rightarrow$C (a) and C$\rightarrow$M (b). The source data are shown by (\protect$\medcircle$) markers and the target data are shown by (\protect$\times$ markers), respectively.  Both the training and testing data are plotted for a one iteration of LOO validation.}
    \label{fig:bda_align}
\end{figure}

Using the TFC in combination with NCA (TFC in Table \ref{tab:het_acc}), led to a significant improvement in generalisation compared to using NCA alone. Moreover, when additional DA was applied generalisation was further improved (TFC+TCA and TFC+BDA in Table \ref{tab:het_acc}). Furthermore, using BDA (TFC+BDA in Table \ref{tab:het_acc}), perfect classification could be achieved in both cases, using a classifier that was trained only using source labels and resulted in features with a low JMMD. This result shows that comparing features that correspond to similar damage-sensitive modes can increase the likelihood the features' conditional distributions are similar, facilitating transfer via unsupervised DA. \\

For C$\rightarrow$M, TCA led to negative transfer,  whereas BDA improves generalisation in all instances, perhaps suggesting that using pseudo-labels to estimate the JMMD is more robust objective for learning a mapping compared to using the MMD in cases where initial classification accuracy is high. This result may be because using the JMMD-based objective in BDA iteratively reduces the MMD between specific classes, meaning it can further reduce small discrepancies between classes, as suggested by the JMMD in Table \ref{tab:het_acc}. \\

 The subset of features selected via the TFC and aligned using NCA are visualised in Figure \ref{fig:PCA_select}, using PCA. In this feature space, mass-states are in close correspondence with their respective state between domains and the features are discriminative. However, the minor-damage classes (Damage 1 and Damage 2) are close; thus, a small shift in the target led to a drop in classification performance, shown in the confusion matrices given in Figure \ref{fig:confuseee}. This result motivates additional DA to further reduce these discrepancies. \\

 Applying additional DA, specifically BDA, successfully reduced this shift in both tasks, resulting in perfect classification (shown in Table \ref{tab:het_acc}). This methodology resulted in a two-dimensional feature space, down from a high-dimensional raw FRF features. An example of the BDA features for training and testing data for both transfer tasks is presented in Figure \ref{fig:bda_align}. It should be noted, a potential limitation of BDA is that it assumes that the label space is homogeneous, which may not always be the case in realistic scenarios; this issue requires further research into partial-DA algorithms \cite{Cao2018a}. \\
 
\begin{figure}[h!]
    \centering
    \begin{subfigure}[b]{0.4\textwidth}
         \centering
        \includegraphics[width=\textwidth]{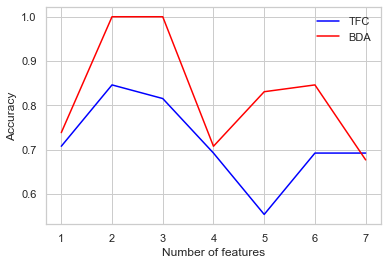}
         \caption{}
         \label{fig:hetro}
     \end{subfigure}
    \begin{subfigure}[b]{0.4\textwidth}
         \centering
         \includegraphics[width=\textwidth]{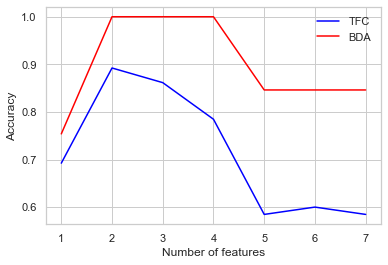}
         \caption{}
         \label{fig:hetro}
     \end{subfigure}
    \caption{Mean accuracy on the target test data for selecting a varying number of features, with (a) representing the transfer from metal to composite (M$\rightarrow$C) and (b) representing the transfer from composite to metal (C$\rightarrow$M). }
    \label{fig:feat_sens2}
\end{figure}

The test accuracy for varying the number of selected feature ranges using the TFC (shown in blue), as well as the result of using both the TFC and BDA (shown in red), is presented in Figure \ref{fig:feat_sens2}. In this case, cross-validation with the source data selected a suitable number of features. It was found that increasing the number of selected features beyond two resulted in worse classification. Although there exist other similar modes, such as Modes One to Three, these modes exhibit less sensitivity to damage, hence their inclusion leads to a negative impact on performance. \\

\begin{figure}[htbp]
  \centering
  \begin{tabular}{cc}
    \includegraphics[width=0.42\textwidth]{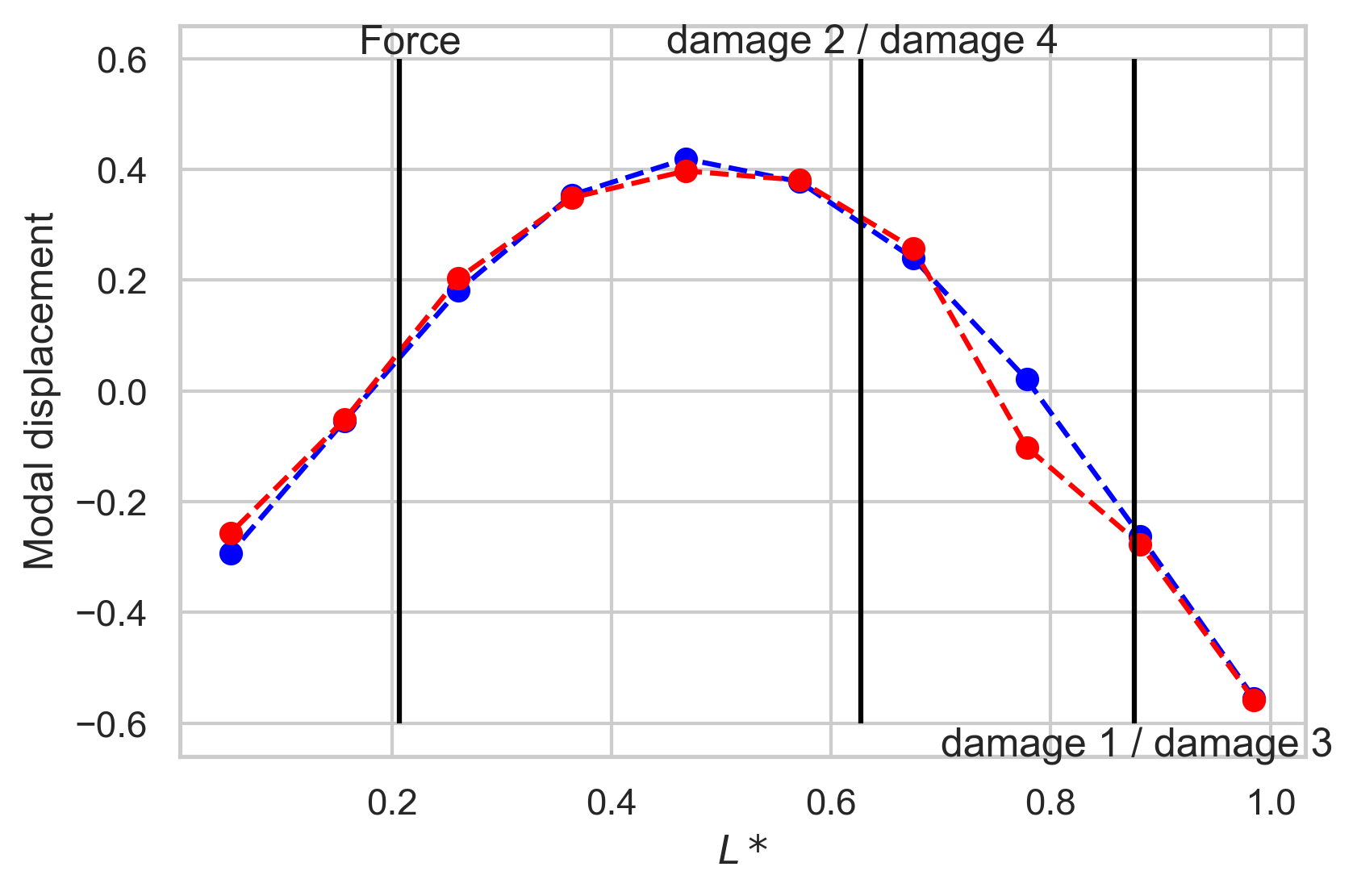} &
    \includegraphics[width=0.42\textwidth]{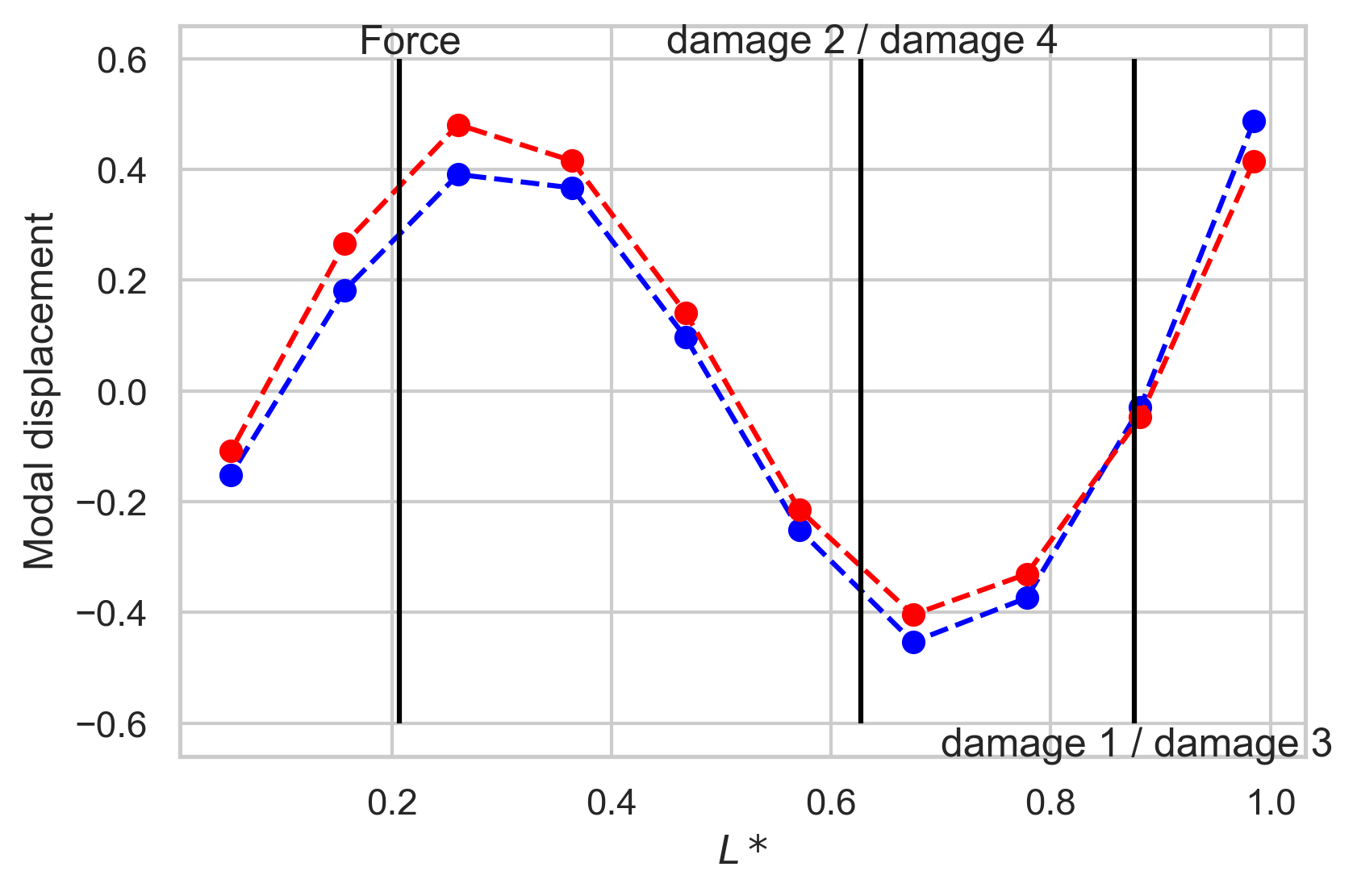} \\
    Mode 1 & Mode 2 \\
    \includegraphics[width=0.42\textwidth]{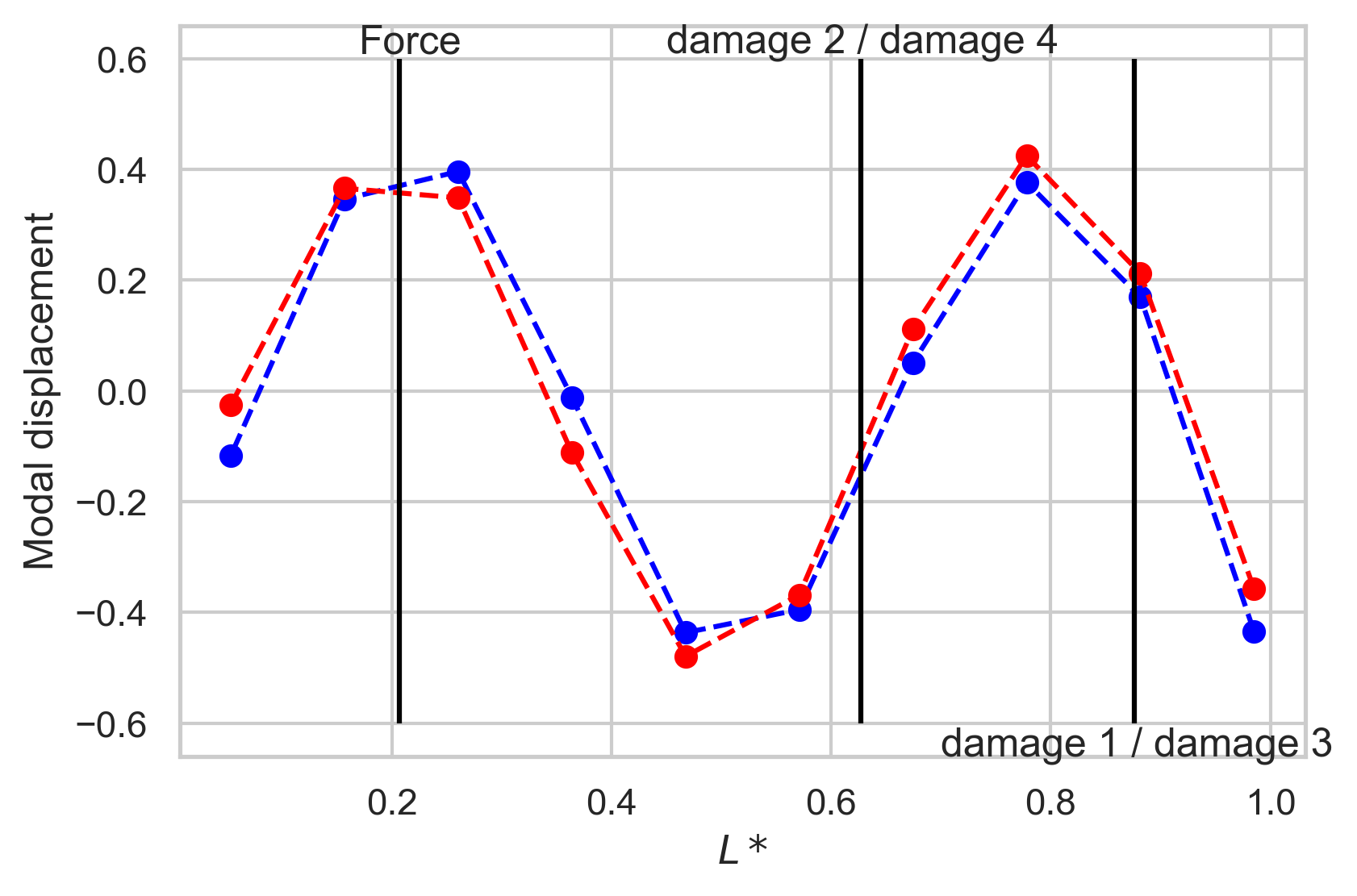} &
    \includegraphics[width=0.42\textwidth]{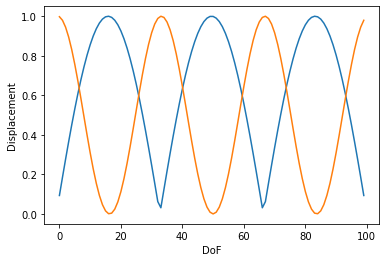} \\
    Mode 3 & Mode 4 \\
    \includegraphics[width=0.42\textwidth]{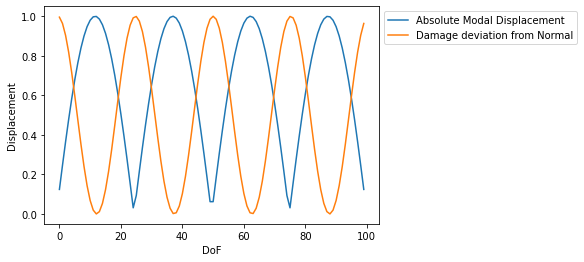} &
    \includegraphics[width=0.42\textwidth]{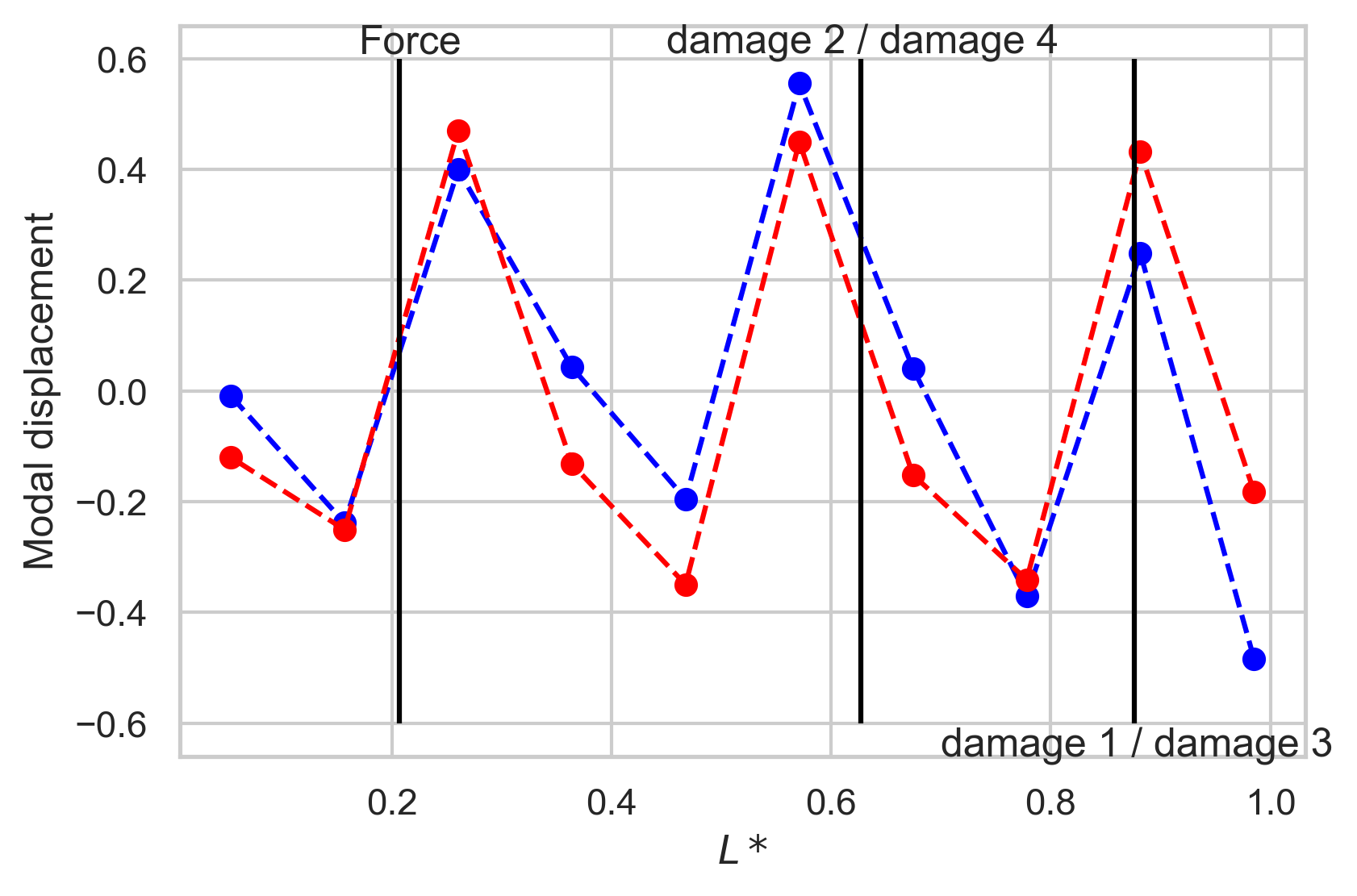} \\
    Mode 5 & Mode 6 \\
    \includegraphics[width=0.42\textwidth]{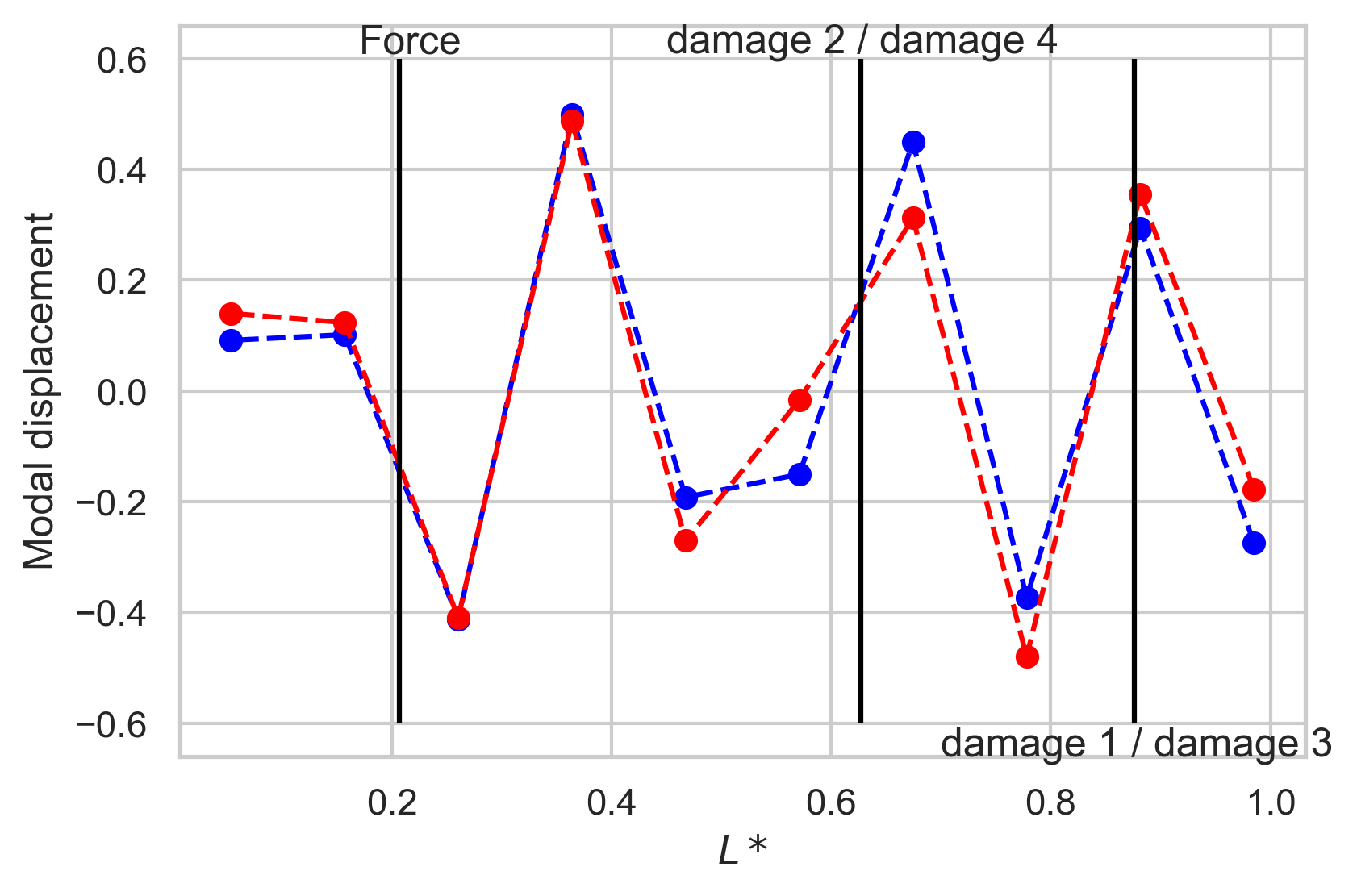} &
    \includegraphics[width=0.42\textwidth]{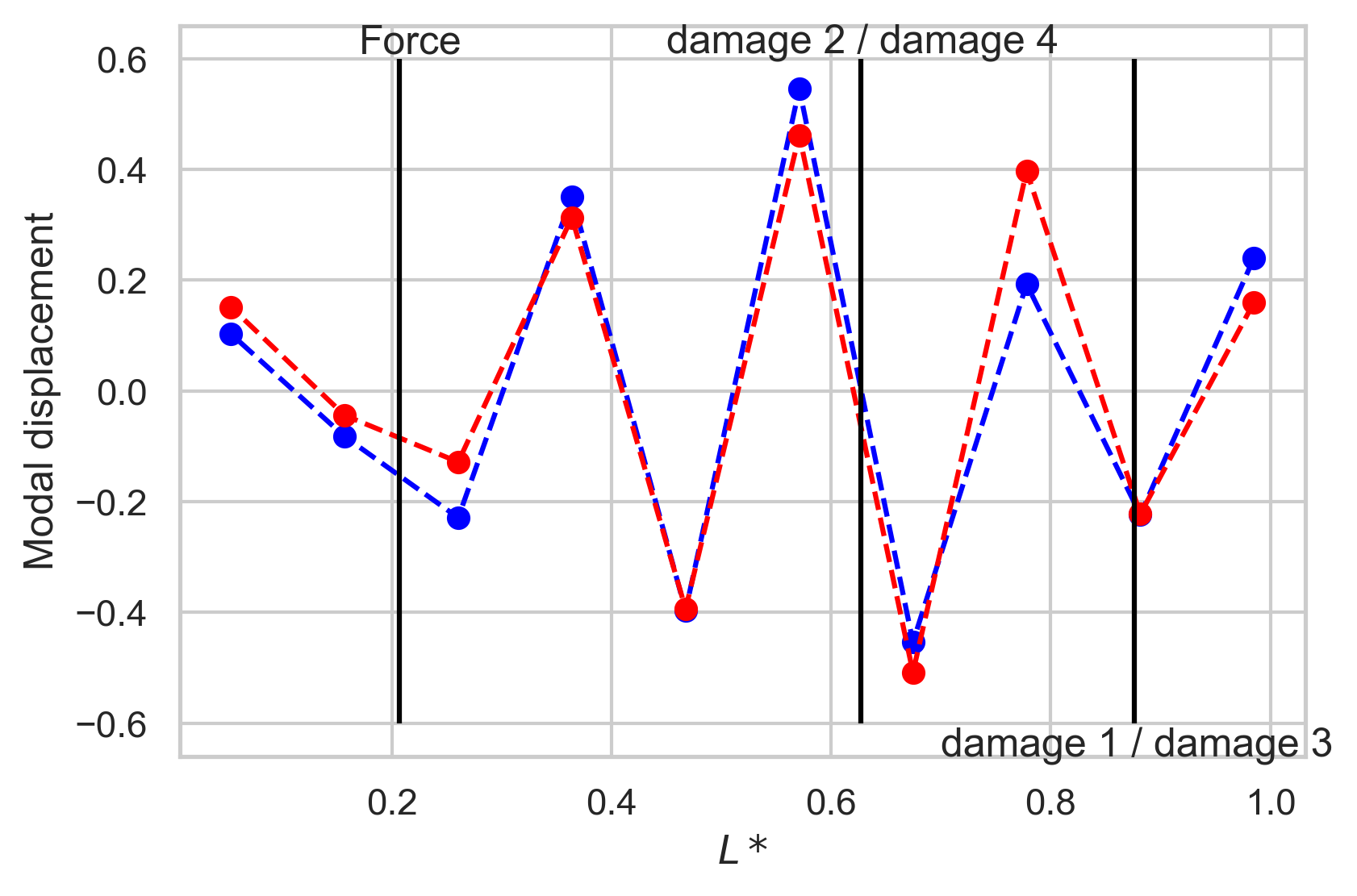} \\
    Mode 7 & Mode 8 \\
  \end{tabular}
  \caption{The first eight identified mode shapes for the metal blade (blue) and composite blade (red). The sensor locations are shown by $\bullet$, and the forcing location, as well as the mass locations, are indicated by the vertical lines.}
  \label{fig:modes}
\end{figure}

 The mode shapes are visualised in Figure  \ref{fig:modes}. The TFC-selected frequencies corresponding to the fourth and fifth modes, and the nodal patterns show a clear similarity in these modes. In addition, masses were located at anti-nodes in Mode Four, and Damage 1 and Damage 3 were located at an anti-node in Mode Five, suggesting these modes would be sensitive to changes in masses in these locations. This figure verifies that the TFC was able to effectively select modes that would be expected to be both sensitive to the pseudo-damage investigated in this study, and similar between domains.\\

\vspace{-0.85cm}

\section{Discussions and conclusion}

Unsupervised TL in a PBSHM framework has the potential to reduce costs and facilitate more in-depth diagnostics for SHM systems. However, these methods require data from different structures to have similar underlying distributions, such that a shared feature space can be found in the absence of target labels. To ensure the distributions are similar, structures, and their corresponding features, must have a similar response to the damage-states of interest.\\

An initial motivating case study demonstrated that unsupervised data-based similarity measures (the PAD and the MMD) are not necessarily indicative of joint distribution similarity for the change of a natural frequency given damage in a discrete location. However, the MAC between the healthy mode shapes was shown to be strongly related to a supervised divergence measure that directly measures the joint distribution similarity -- the JMMD. Thus, a feature selection criterion incorporating the MAC was proposed to identify sets of features that meet the conditional distribution similarity assumption required by unsupervised TL methods. By selecting a subset of features the assumption that all features are strongly related is relaxed, allowing for this assumption to hold for only a subset of features \cite{rojas2018invariant}. Furthermore, this measure only requires data from the undamaged structure, making it applicable to a wide range of PBSHM scenarios.\\

The proposed feature selection criterion was validated via two case studies. The first involved a heterogeneous numerical population with 342 transfer tasks (transfer of a damage classifier). Using the MAC to select a subset of related features (natural frequencies) led to significantly improved generalisation across the population. In contrast, two popular nonlinear DA algorithms (TCA and BDA) did not improve upon a linear transform  (learned via NCA), on average across the population. This study showed that, for a diverse population, selecting features with similar undamaged mode shapes can greatly enhance generalisation between source and target domains using unsupervised DA (NCA), with minimal risk of negative transfer. Moreover, after applying the TFC, BDA improved generalisation further, suggesting it can be used to select suitable features for use with conventional DA methods. These future studies should also include various realistic damage modes to investigate when certain damage labels can be transferred.\\

The second case study investigated transferring a damage classifier to predict different damage locations and extents in a heterogeneous pair of helicopter blades. This case study highlights aspects regarding the importance of selecting related features. FRF frequencies were used as features. As an initial processing step, frequencies that had a high contribution from the natural frequencies were put in correspondence across the source and target domains. Using the proposed feature criterion in conjunction with unsupervised DA (NCA and BDA), a set of features corresponding to similar, and damage-sensitive modes could be selected, and a low-dimensional feature space could be inferred, resulting in perfect classification in the target domain using a classifier trained using only source data for both transfer tasks. Thus, this case study demonstrates a methodology, combining prior knowledge about the physical behaviour of frequency-based features, with unsupervised DA methods, as capable of extracting a shared low-dimensional feature space from high-dimensional frequency data.\\

The proposed method is a step towards demonstrating how engineering knowledge can be used to inform what features can be transferred. Nevertheless, there are several potential issues with using the MAC, and future work is required to extend the findings in this paper. First, the requirements for the sensor networks should be investigated, considering both optimal sensor locations for individual structures and identifying corresponding sensor locations between structures. Previous work has focussed on selecting informative subsets of sensors \cite{worden2004overview}, which could be extended to considering sensors across multiple structures. Another interesting approach would be to interpolate between sensors, either using FE or statistical models. In addition, this approach should be evaluated in structures under realistic operating conditions, to evaluate robustness to noisy identification of the modes, the influence of environmental and operating conditions, and nonlinearity.\\

This paper highlights the importance of selecting features with similar mode shapes for effective transfer. While lower modes are more likely to be similar across structures, they tend to be less sensitive to damage. In contrast, the most discriminative vibration-based features for damage localisation often correspond to local modes, which are harder to consistently identify across different structures. Further research should investigate which damage identification tasks can feasibly transfer using vibration-based features. Future work could also consider transferring different features that are more sensitive to certain types damage, and developing associated criteria to select transferable features.

\section*{Acknowledgements}

The authors would like to acknowledge the support of the UK Engineering and Physical Sciences Research Council via grants EP/R006768/1, EP/R004900/19 and EP/W005816/1. For the purpose of open access, the authors have applied for a Creative Commons Attribution (CC-BY-ND) licence to any Author Accepted Manuscript version arising.

\newpage

\begin{appendices}

\section{List of acronyms}

\begin{table}[h]
\centering
\caption{A summary of the acronyms used throughout this paper.}
\begin{tabular}{ll}
\hline
\textbf{Acronym} & \textbf{Definition} \\ \hline
PBSHM            & Population-based structural health monitoring \\
DA               & Domain adaptation \\
TL               & Transfer learning \\
SHM              & Structural health monitoring \\
MAC              & Modal assurance criterion \\
MMD              & Maximum mean discrepancy \cite{Gretton2012}\\
PAD              & Proxy-a distance \cite{Ben-David2007}\\
JMMD             & Joint maximum mean discrepancy \cite{Long2013}\\
SVM              & Support vector machine \\
RBF              & Radial basis function \\
NCA              & Normal condition alignment \cite{poole2022statistic} \\
KNN              & K-nearest neighbours \\
RKHS             & Reproducing kernel Hilbert space \\
LOO              & Leave-one-out (validation) \\
TCA              & Transfer component analysis \cite{Pan2010} \\
TFC              & Transfer feature criterion \\
BDA              & Balanced distribution adaptation \cite{Wang2017}\\

 \hline
\end{tabular}
\label{tab:acronyms}
\end{table}

\section{Motivating example: relationship between mode shapes and damage}

\begin{figure}[h!]
    \centering
    \begin{subfigure}[b]{0.48\textwidth}
         \centering
        \includegraphics[width=\textwidth]{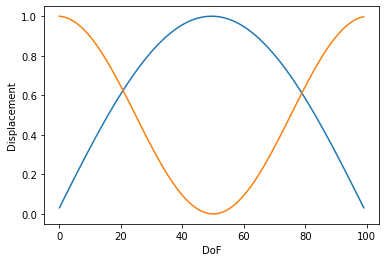}
         \caption{}
         \label{fig:hetro}
     \end{subfigure}
    \begin{subfigure}[b]{0.48\textwidth}
         \centering
         \includegraphics[width=\textwidth]{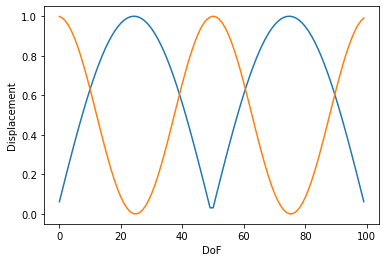}
         \caption{}
         \label{fig:hetro}
     \end{subfigure}
    \caption{A comparison between the (absolute) mode shape of an undamaged 100DoF chain of masses (blue) and the relative discrepancy in the associated natural frequency caused by a stiffness reduction at each DoF (orange) for the first (a) and second mode (b); both are normalised between 0 and 1.}
    \label{fig:shapez}
\end{figure}

To illustrate this relationship, a 100 degree of freedom (DoF) chain of masses connected with springs and dampers was considered. The mode shapes and natural frequencies were found via the eigenvalue problem for the undamaged structure, as well as the natural frequencies for a given stiffness reduction (damage) at each DoF. The absolute value of the mode shapes was compared to the difference between damaged and undamaged natural frequency ($\omega_{d,normal}-\omega_{d,damage}$), each normalised such that $x \in [0,1]$. The comparison of the first two modes (blue) and the scaled discrepancy of the associated natural frequencies (orange) are shown in Figure \ref{fig:shapez}. It can be seen the deviation of the natural frequencies is inversely related to the modal displacement, suggesting that they could provide an indication of which damage locations will correspond between structures\footnote{Only the natural frequencies are considered in this paper, but more generally the mode shapes indicate areas of sensitivity to damage for each mode, so may also be indicative of similarity of other vibration-based features.}. This example shows the scaled discrepancy; thus, it suggests that if the modal displacement for a given location corresponds between structures, the only differences will be caused by the scale of the features and differences in damage extent; the former problem can be addressed via unsupervised TL \cite{poole2022statistic}. \\

\section{Non-dimensionalisation of corresponding modes for transfer}

To transfer damage labels, it is important that features can be designed such that the conditional distributions are in close correspondence. Consider the problem of modelling the conditional distribution of damage location $l$ and extent $\delta$, using natural frequencies as damage-sensitive features. It must hold that,

\begin{equation}
    p(l, \delta | \phi_s(\boldsymbol{\omega_s})) \approx  p(l, \delta | \phi_t(\boldsymbol{\omega_t}))
\end{equation}

\noindent where $\phi_s$ and $\phi_t$ represent a source and target feature map. If the systems are assumed to be linear, the generative process for the features is well understood. Consider a system as a set of lumped masses -- for ease of explanation, damping is ignored -- the natural frequencies can be found by, 

\begin{equation}
    \omega_i = \sqrt{\frac{\hat{k}_i}{\hat{m}_i}}
\end{equation}

\noindent where $\hat{k}_i$ and $\hat{m}_i$ are the modal stiffness and mass for the $i^{\text{th}}$ mode. In reality, $\omega_i$ should be considered as a random variable, with variation caused by several factors, such as EoVs and acquisition noise, but for demonstration purposes, this deterministic value is used \footnote{This equation could be considered to be modelling the mean under ideal circumstances.}. These values are related to the geometry and material properties of the system by,
\begin{equation}
\begin{aligned}
    \hat{k}_i = \boldsymbol{\psi}^{(i)^T} K   \boldsymbol{\psi}^{(i)}\\ \hat{m}_i = \boldsymbol{\psi}^{(i)^T} M   \boldsymbol{\psi}^{(i)}
\end{aligned}
\end{equation}
where $K$ and $M$ are the stiffness and mass matrices respectively. From these equations, it is apparent, given that the structures have the same geometry, and are uniform, differences in stiffness and mass can be removed by scaling the features:

\begin{equation}
\begin{aligned}   \frac{\boldsymbol{\psi}^{(i)^T}_{s} K_s   \boldsymbol{\psi}^{(i)}_{s}}{k_s} = \frac{\boldsymbol{\psi}^{(i)^T}_{t} K_t   \boldsymbol{\psi}^{(i)}_{t}}{k_t} \\ \frac{\boldsymbol{\psi}^{(i)^T}_{s} M_s   \boldsymbol{\psi}^{(i)}_{s}}{m_s} = \frac{\boldsymbol{\psi}^{(i)^T}_{t} M_t   \boldsymbol{\psi}^{(i)}_{t}}{m_t}
\end{aligned}
\end{equation}
where the subscripts $s$ and $t$ refer to source and target parameters respectively, and $k$ and $m$ are the stiffness and mass values. After scaling, a percentage reduction in stiffness in a spring $\alpha = \frac{k_i}{k}$ will correspond in both sets of features. This scaling could be inferred from data and/or physics; for example, the tip stiffness of a cantilever beam is given by $k = \frac{EI}{l^3}$, which could be estimated with knowledge of the structure. \\

This equation establishes a basis for transfer between structures which mainly differ by a global parameter (i.e. different Young's Modulus and density), but what if there are also local discrepancies? The mode shapes may also provide insight into which features can be transferred in this case. To corroborate this point, consider the stiffness matrix of a 2DoF system fixed at both ends,
\begin{equation}
    K = \begin{bmatrix}
    k_{1} + k_{2}       & -k_{2} \\
    -k_{2}       & k_{2} + k_{3} 
\end{bmatrix}
\end{equation}
By using eq(13), this can be decoupled in the modal stiffness as a function of the mode shapes, 
\begin{equation}
    \hat{k}_i = \psi^{(i)^2}_{1} (k_1 + k_2) + \psi^{(i)^2}_{2} (k_2 + k_3) - \psi^{(i)}_{1}\psi^{(i)}_{2}k_2
\end{equation}
In this example, if the displacement in the first and second DoF is equivalent $\psi_{1,i}=\psi_{2,i}$, which would give an anti-node between the two masses, the $k_2$ terms cancel, meaning any discrepancies between the structures at this location would not affect this mode, so assuming $k_1$ and $k_3$ were similar, corresponding features should have related behaviour. While this is a simple example, in principle this should scale to larger, more complex structures to some extent and it generally suggests that if modal displacement between two locations is small, discrepancies will only cause small changes to the response between structures.\\

This short analysis lays out two core challenges for transferring labelled  frequency data generated by the linear response of a structure:
\begin{enumerate}
    \item Features should be selected such that they can be non-dimensionalised to bring features into alignment; this issue is approached by using the mode shapes and unsupervised domain adaptation in this paper.
    \item A label mapping will be required in some cases to equate damage to an extent (i.e. a percentage reduction stiffness). For example, a label may detail the length of a crack, but if the geometry of the structures is dissimilar the same crack would result in different relative reductions in stiffness and therefore frequency. An initial attempt to address this issue could include scaling the label to correspond to the same reduction based on analytical or FE models. 
\end{enumerate}

\section{Case study: finite-element beams}

\begin{figure}[h!]
    \centering
    \begin{subfigure}[b]{0.5\textwidth}
         \centering        \includegraphics[width=\textwidth]{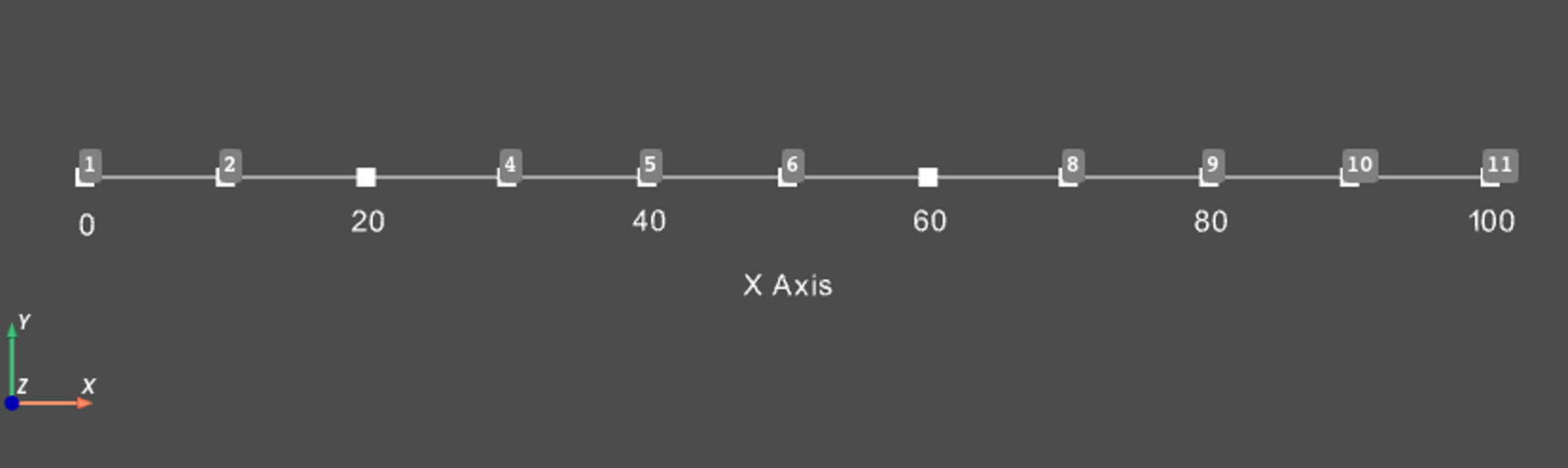}
         \caption{}
         \label{fig:hetro}
     \end{subfigure}
    \begin{subfigure}[b]{0.5\textwidth}
         \centering
         \includegraphics[width=\textwidth]{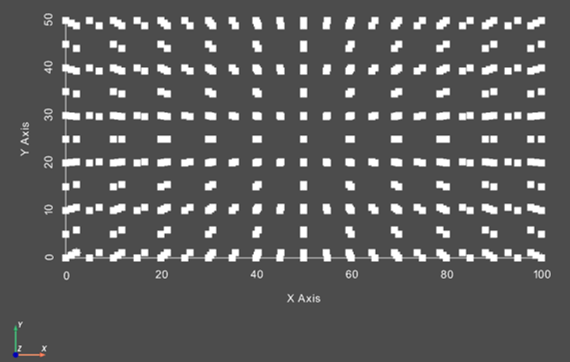}
         \caption{}
         \label{fig:hetro}
     \end{subfigure}
     \caption{Node plots for the 2D and 3D FE models of a beam.}
      \label{fig:macFE}
     \end{figure}

To demonstrate the idea of using mode shapes to select similar damage-sensitive features, an example consisting of two beams generated via finite-element analysis (FEA) is presented. The task is to transfer between a simple one-dimensional beam, simulated using Euler-Timoshenko elements, to a beam simulated with three-dimensional solid elements with the same geometry; the node plots are shown in Figures \ref{fig:macFE}a and \ref{fig:macFE}b respectively. Both beams were simulated with a length of 1m, width of 0.5m and thickness of 0.1m. \\

\begin{figure}[h!]
    \centering
    \begin{subfigure}[b]{0.32\textwidth}
         \centering        \includegraphics[width=\textwidth]{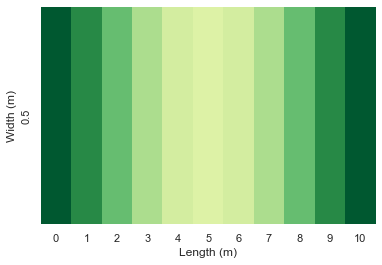}
         \caption{}
         \label{fig:hetro}
     \end{subfigure}
    \begin{subfigure}[b]{0.32\textwidth}
         \centering
    \includegraphics[width=\textwidth]{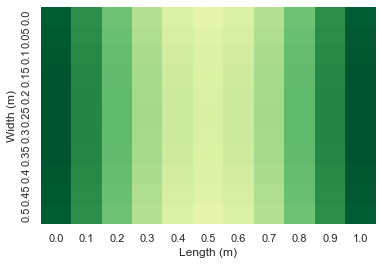}
         \caption{}
         \label{fig:hetro}
     \end{subfigure}
     \begin{subfigure}[b]{0.32\textwidth}
         \centering
        \includegraphics[width=\textwidth]{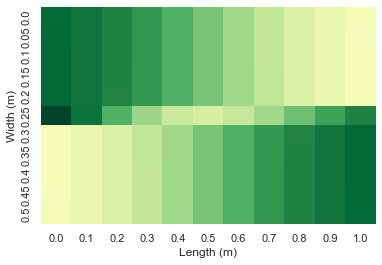}
         \caption{}
         \label{fig:hetro}
     \end{subfigure}
    \caption{The first bending mode in the 1D (a) and 3D FE beam (b), as well as a torsional mode in the 3D beam (c) in the z-axis.}
    \label{fig:mode_shapes}
\end{figure}

This case presents a number of interesting considerations for transfer using the mode shapes. Firstly, variation between the two beams will be present because the three-dimensional beam can simulate varying displacement across the thickness and width of the beam, whereas the one-dimensional beam cannot; thus, only pure bending modes will be in direct correspondence. The first bending mode (in the $z$-direction) of the source and target is given in Figures \ref{fig:mode_shapes}a and Figure \ref{fig:mode_shapes}b respectively, which are in correspondence, as well as the first torsional mode in the target (in the $z$-direction); this mode cannot be simulated by the source. In addition, the three-dimensional beam contains a much larger number of elements (sensing locations), so direct comparison of the mode shapes is challenging; this could be a prominent issue when transferring between two real structures with different sensor arrays. Here, the line of nodes along the centre line is considered, which should be able to identify pure bending modes.\\

In addition, a mismatch in elements presents a problem where the label spaces are heterogeneous, as damage can not be identified across the width of the one-dimensional beam; however, it is hypothesised that such a model could still be used to locate damage along the length of the beams by using bending modes. As such, the transfer task considered is a damage-localisation task, where damage was simulated as a 5\% reduction of Young's Modulus in the elements along the width at $0.25L$, $0.5L$ and, $0.75L$, where $L$ is the length of the beam. Fifty samples for the undamaged beams and each damage class were simulated, with variation being introduced by adding Gaussian noise to the Young's Modulus; the beams were assumed to have the material properties of steel, with a Young's modulus of 210GPa and density of 7800kg/$m^3$.\\

     \begin{figure}
     \begin{subfigure}[b]{0.32\textwidth}
         \centering
         \includegraphics[width=\textwidth]{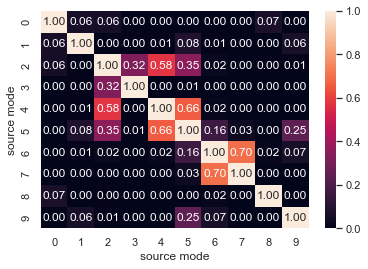}
         \caption{}
         \label{fig:hetro}
     \end{subfigure}
     \begin{subfigure}[b]{0.32\textwidth}
         \centering
         \includegraphics[width=\textwidth]{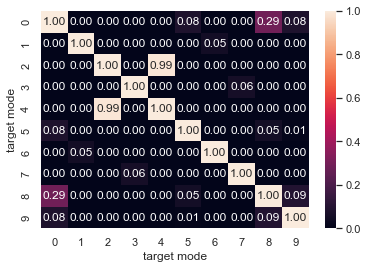}
         \caption{}
         \label{fig:hetro}
     \end{subfigure}
     \begin{subfigure}[b]{0.32\textwidth}
         \centering
         \includegraphics[width=\textwidth]{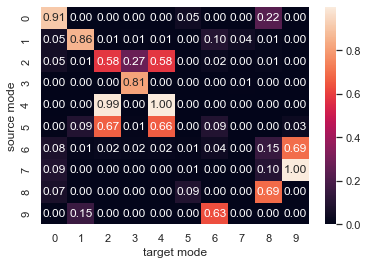}
         \caption{}
         \label{fig:hetro}
     \end{subfigure}
    \caption{The MAC matrix between the first ten modes identified from the 1D (a) and 3D beams (b), using a subset of nodes along the centre line for the 3D beam, and between the 1D and 3D beams (c).}
    \label{fig: macFE}
\end{figure}

\begin{figure}[h!]

     \centering        
     \includegraphics[width=0.5\textwidth]{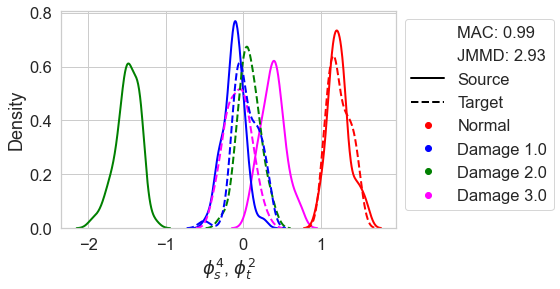}
     
    \caption{A KDE plot comparing the natural frequencies of the second source mode with the fourth target mode, demonstrating the problem of comparing mode shapes with limited sensor networks. Density is normalised for each class independently.}
    \label{fig: crry}
\end{figure}

The mode shapes in the $x$,$y$ and $z$ coordinates were concatenated and the MAC matrices for the source (1D beam), target (3D beam), as well as, between the source and target are presented in Figures \ref{fig: macFE}a, \ref{fig: macFE}b, and  \ref{fig: macFE}c respectively. Initially, it can be seen that two modes in the target are identified as almost identical (Mode 2 and Mode 4), which is because target Mode 2 is predominantly a torsional Mode in $z$, whereas Mode 4 is the first bending Mode in $y$, but as only the centre line nodes are considered, the contribution of this torsional mode is negligible in the MAC, as displacement in the centre is small. Here, potential misidentification of the mode shapes because of insufficient sensor resolution illustrates a major challenge for selecting domain-invariant features, as utilising source Mode 4 and target Mode 2 would introduce a dissimilar feature into the set, potentially causing negative transfer, as shown by the kernel density plot (KDE; the interested reader may refer to \cite{Murphy2014}), of the natural frequencies in Figure \ref{fig: crry}.\\

\begin{figure}[h!]
    \centering
    \begin{subfigure}[b]{0.29\textwidth}
         \centering        \includegraphics[width=\textwidth]{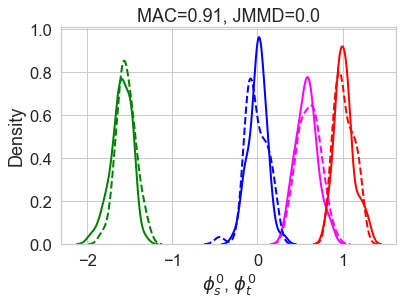}
         \caption{}
         \label{fig:hetro}
     \end{subfigure}
     \begin{subfigure}[b]{0.29\textwidth}
         \centering        \includegraphics[width=\textwidth]{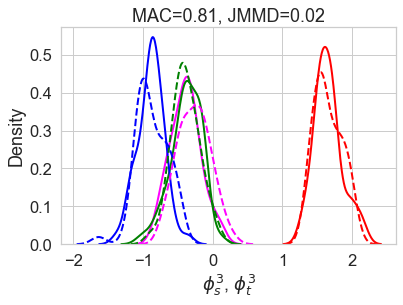}
         \caption{}
         \label{fig:hetro}
     \end{subfigure}
     \begin{subfigure}[b]{0.29\textwidth}
         \centering        \includegraphics[width=\textwidth]{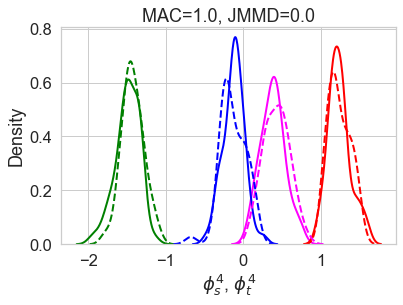}
         \caption{}
         \label{fig:hetro}
     \end{subfigure}
    \begin{subfigure}[b]{0.29\textwidth}
         \centering        \includegraphics[width=\textwidth]{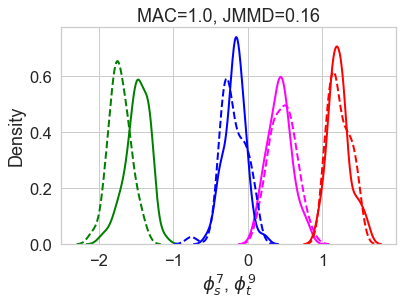}
         \caption{}
         \label{fig:hetro}
     \end{subfigure}
    \begin{subfigure}[b]{0.41\textwidth}
         \centering
      \includegraphics[width=\textwidth]{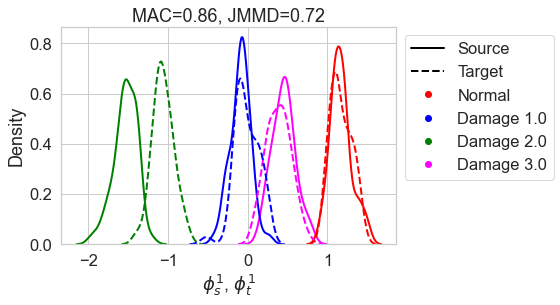}
         \caption{}
         \label{fig:hetro}
     \end{subfigure}
    
    \caption{KDE plots of the TFC selected features for a beam generated as a 1D chain of beam elements (source) and a beam generated with a 3D geometry (target). Density is normalised for each class independently.}
    \label{fig: kde1}
\end{figure}

The TFC was applied to extract the related features; here five features were selected, as there are five independent pairs of modes with MAC values above 0.8. In addition, NCA was applied to reduce distribution shift. The selected features are visualised using KDE plots, as shown in Figure \ref{fig: kde1}. By selecting corresponding bending modes, the associated natural frequencies follow similar joint distributions, shown by the near-zero JMMD values. The MAC discrepancy, accuracy for damage localisation and JMMD, for all the features, TFC-selected features and the features not selected by the TFC are presented in Table 2. While the original set of features could achieve perfect classification for this task, the JMMD value suggests the data were generated from two distinct distributions, whereas the TFC features have a JMMD near zero, indicating the opposite. Furthermore, the remaining features have a significantly higher JMMD, suggesting the TFC was able to extract the shared information.  \\ 

\begin{table}[h!]
    \centering
     \caption{The MAC discrepancy ($d_{MAC}$), accuracy and JMMD values for the first ten natural frequencies, GA-selected frequencies and the unselected frequencies for the 1D and 3D FE beams.}
    \label{tab: nom}
    \begin{tabular}{llll}
    \toprule
    & $d_{MAC}$ & Accuracy & JMMD \\
    \midrule
    All features & 0.49 & 1.00 & 1.52 \\
    TFC features & 0.92 & 1.00 & 0.15 \\
    Unselected features & 0.07 & 0.63 & 2.94 \\
    \bottomrule
    \end{tabular}
   
\end{table}

\newpage

\section{Experimental case study: heterogeneous blades - additional details}

\begin{table}[h!]
\caption{Table summarising sensor location for the experiments on the blades, where L* is the non-dimensionalised length and W* is the non-dimensionalised width. The final two sensors in the composite blade change in location as the blade gets thinner close to the root.}
\begin{tabular}{|l|l|l|l|l|l|l|l|l|l|l|l|}
\hline
Sensor no. & 1     & 2     & 3     & 4     & 5     & 6     & 7     & 8     & 9     & 10    & Force \\ \hline
L*         & 0.053 & 0.157 & 0.260 & 0.364 & 0.467 & 0.571 & 0.674 & 0.778 & 0.881 & 0.985 & 0.315 \\ \hline
W* metal   & 0.666 & 0.666 & 0.666 & 0.666 & 0.666 & 0.666 & 0.666 & 0.666 & 0.666 & 0.666 &  0.275     \\ \hline
W*  comp   & 0.666 & 0.666 & 0.666 & 0.666 & 0.666 & 0.666 & 0.666 & 0.666 & 0.5 & 0.133 &  0.283     \\ \hline
\end{tabular}
\end{table}

\begin{table}[h!]
\caption{Table presenting the natural frequencies identified for the metal and composite blades, for the normal condition and four mass states.}
\begin{tabular}{l|l|l|l|l|l|l|l|l|}
\cline{2-9}
                                      & $\omega_1$ & $\omega_2$ & $\omega_3$ & $\omega_4$ & $\omega_5$ & $\omega_6$ & $\omega_7$ & $\omega_8$ \\ \specialrule{2.5pt}{1pt}{1pt}
\multicolumn{1}{|l|}{Metal Blade: Normal} & 5.46       & 16.52      & 32.56      & 55.76      & 83.76      & 114.24     & 154.61     & 197.07     \\ \hline
\multicolumn{1}{|l|}{Comp Blade: Normal}  & 4.28       & 12.70      & 26.83      & 45.95      & 69.26      & 96.84      & 127.13     & 160.37     \\ \hline
\multicolumn{1}{|l|}{Ratio: Normal}       & 1.28       & 1.30       & 1.21       & 1.21       & 1.21       & 1.18       & 1.22       & 1.22       \\ \specialrule{2.5pt}{1pt}{1pt}
\multicolumn{1}{|l|}{Metal Blade: Damage 1} & 5.44       & 16.55      & 32.64      & 55.63      & 83.51      & 113.82     & 154.14     & 196.04     \\ \hline
\multicolumn{1}{|l|}{Comp Blade: Damage 1}  & 4.28       & 12.87      & 26.79      & 45.86      & 68.87      & 96.4       & 126.64     & 159.89     \\ \hline
\multicolumn{1}{|l|}{Ratio: Damage 1}       & 1.27       & 1.29       & 1.22       & 1.21       & 1.21       & 1.18       & 1.22       & 1.23       \\ \specialrule{2.5pt}{1pt}{1pt}
\multicolumn{1}{|l|}{Metal Blade: Damage 2} & 5.44       & 16.74      & 32.65      & 55.47      & 83.68      & 113.77     & 153.90     & 196.01     \\ \hline
\multicolumn{1}{|l|}{Comp Blade: Damage 2}  & 4.24       & 12.85      & 26.81      & 45.75      & 69.09      & 96.52      & 126.36     & 160.04     \\ \hline
\multicolumn{1}{|l|}{Ratio: Damage 2}       & 1.28       & 1.30       & 1.22       & 1.21       & 1.21       & 1.18       & 1.22       & 1.22       \\ \specialrule{2.5pt}{1pt}{1pt}
\multicolumn{1}{|l|}{Metal Blade: Damage 3} & 5.46       & 16.52      & 32.49      & 55.44      & 82.97      & 113.55     & 154.03     & 195.77     \\ \hline
\multicolumn{1}{|l|}{Comp Blade: Damage 3}  & 4.28       & 12.72      & 26.73      & 45.64      & 68.28      & 98.09      & 126.6      & 160.01     \\ \hline
\multicolumn{1}{|l|}{Ratio: Damage 3}       & 1.28       & 1.3        & 1.22       & 1.21       & 1.22       & 1.16       & 1.22       & 1.22       \\ \specialrule{2.5pt}{1pt}{1pt}
\multicolumn{1}{|l|}{Metal Blade: Damage 4} & 5.47       & 16.31      & 32.53      & 54.92      & 83.54      & 112.74     & 152.58     & 196.06     \\ \hline
\multicolumn{1}{|l|}{Comp Blade: Damage 4}  & 4.24       & 12.67      & 26.81      & 45.30      & 68.89      & 97.15      & 125.74     & 160.29     \\ \hline
\multicolumn{1}{|l|}{Ratio: Damage 4}       & 1.29       & 1.29       & 1.21       & 1.21       & 1.21       & 1.16       & 1.21       & 1.22       \\ \hline
\end{tabular}
\end{table}

\begin{table}[h!]
\centering
\caption{Table providing the sequence of experiments on the metal and composite blades.}
\begin{tabular}{|l|l|l|}
\hline
Test no. & Mass state & Repeats \\ \hline
1        & Normal         & 15      \\ \hline
2        & Damage 3        & 10      \\ \hline
3        & Normal         & 3       \\ \hline
4        & Damage 4         & 10      \\ \hline
5        & Normal         & 3       \\ \hline
6        & Damage 1         & 10      \\ \hline
7        & Normal         & 1       \\ \hline
8        & Damage 2         & 10      \\ \hline
9        & Normal         & 3       \\ \hline
\end{tabular}
\end{table}

\newpage

\end{appendices}
\bibliographystyle{IEEEtran}
\bibliography{ref}
\newpage

\pagenumbering{roman}

\end{document}